%% file: main.tex
\documentclass[10pt,twocolumn,letterpaper]{article}

\usepackage{iccv}              % To produce the CAMERA-READY version
\input{preamble}

\definecolor{iccvblue}{rgb}{0.21,0.49,0.74}
\usepackage[pagebackref,breaklinks,colorlinks,allcolors=iccvblue]{hyperref}
\usepackage{array,multirow,graphicx}
\usepackage{rotating}
\usepackage{arydshln}
\usepackage{mathtools}
\usepackage{footmisc}
\usepackage[accsupp]{axessibility}  % Improves PDF readability for those with disabilities.

\def\paperID{****} % *** Enter the Paper ID here
\def\confName{ICCV}
\def\confYear{2025}

\title{Scene Coordinate Reconstruction Priors}

\author{
Wenjing Bian\textsuperscript{2,*}\quad
Axel Barroso-Laguna\textsuperscript{1}\quad
Tommaso Cavallari\textsuperscript{1} \quad \\
Victor Adrian Prisacariu\textsuperscript{1,2} \quad
Eric Brachmann\textsuperscript{1}\\
\textsuperscript{1}Niantic Spatial \quad \textsuperscript{2}University of Oxford\\\\
\url{https://nianticspatial.github.io/scr-priors/}}

\begin{document}

\input{main_content}    

{
    \small
    \bibliographystyle{ieeenat_fullname}
    \bibliography{main}
}

\input{supplement}
\end{document}

%% file: preamble.tex
\newcommand\blfootnote[1]{%
  \begingroup
  \renewcommand\thefootnote{}\footnote{#1}%
  \addtocounter{footnote}{-1}%
  \endgroup
}

%% file: main_content.tex
\twocolumn[{%
\renewcommand\twocolumn[1][]{#1}%
\maketitle
\begin{center}
    \centering
    \captionsetup{type=figure}
    \vspace{-1cm}
    \includegraphics[width=\textwidth]{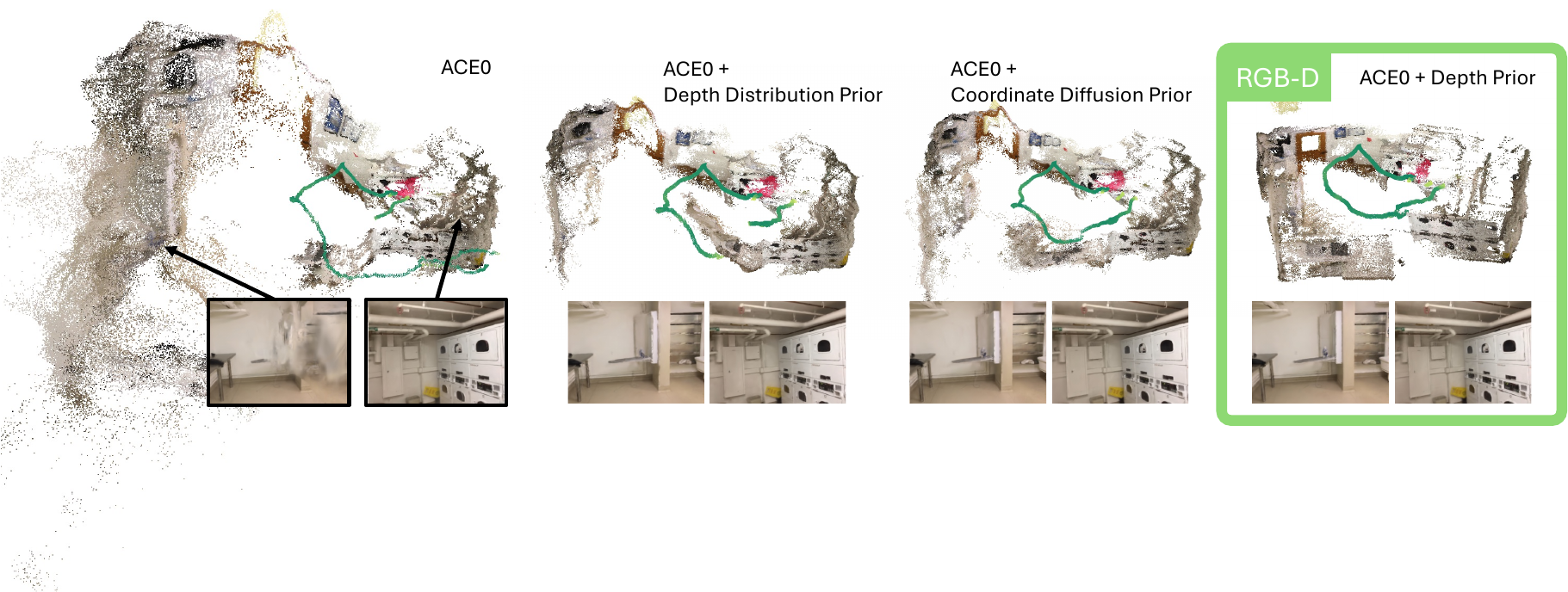}
    \vspace{-2.4cm}
    \captionof{figure}{\textbf{Reconstruction Priors.} ACE0 \cite{brachmann2024acezero}, a neural SfM pipeline, struggles with an indoor scene (left). The reconstruction partly degenerates, as is evident by the dispersed point cloud, camera poses (green) penetrating a wall, and artifacts in one of the views synthesized with Splatfacto \cite{nerfstudio} (bottom). We incorporate various priors into ACE0 which lead to a more consistent scene layout, better pose estimates, and fewer artifacts in synthesized views (center). We also show a version of ACE0 taking RGB-D rather than RGB images as input (right).}
    \label{fig:teaser}
    \vspace{0.3cm}
\end{center}%
}]
\maketitle
\blfootnote{* Work done during an internship at Niantic.}

\begin{abstract}
Scene coordinate regression (SCR) models have proven to be powerful implicit scene representations for 3D vision, enabling visual relocalization and structure-from-motion. 
SCR models are trained specifically for one scene.
If training images imply insufficient multi-view constraints SCR models degenerate.
We present a probabilistic reinterpretation of training SCR models, which allows us to infuse high-level reconstruction priors.
We investigate multiple such priors, ranging from simple priors over the distribution of reconstructed depth values to learned priors over plausible scene coordinate configurations.
For the latter, we train a 3D point cloud diffusion model on a large corpus of indoor scans.
Our priors push predicted 3D scene points towards plausible geometry at each training step to increase their likelihood.
On three indoor datasets our priors help learning better scene representations, resulting in more coherent scene point clouds, higher registration rates and better camera poses, with a positive effect on down-stream tasks such as novel view synthesis and camera relocalization.
\end{abstract}

\section{Introduction}

With the recent advent of learning-based structure-from-motion (SfM) pipelines, there are three avenues to further push their capabilities: 1) priors, 2) priors, and 3) priors.

In this work, we consider a particular class of neural SfM models that hitherto only incorporated very weak priors: Scene Coordinate Regression (SCR) \cite{shotton2013scene}.
SCR models are implicit scene representations. 
They are trained on images of a particular scene with known ground truth camera poses \cite{shotton2013scene, brachmann2016, Brachmann2018dsacpp, Brachmann2021dsacstar, brachmann2023ace}.
Once trained, SCR models allow to estimate the camera poses of unseen query images relative to the training scene, \ie they enable visual relocalization.
Recently, ACE0 \cite{brachmann2024acezero} showed self-supervised training of SCR, turning it into a fully differentiable, neural SfM approach.

SCR reinterprets classical 3D reconstruction using a machine learning methodology.
However, they still follow classical 3D vision principles when learning the 3D geometry of a scene:
where feature-based methods triangulate sparse key points explicitly, SCR methods rely on dense, implicit triangulation of image patches \cite{Brachmann2018dsacpp}.
Whether one triangulates implicitly or explicitly, without sufficient multi-view constrains triangulation fails \cite{hartley2003multiple}.
This happens \eg for texture-poor areas, repetitive structures, reflections, \etc where multi-view observations cannot be resolved to the same 3D scene point.
In Fig.~\ref{fig:teaser}, featureless walls and floor cause ACE0 to degenerate with an adverse effect on novel view synthesis on top of the ACE0 reconstruction.

Our aim is to overcome the ambiguity and potential incompleteness of scene-specific training data, by leveraging high-level reconstruction priors.
It is easy to see that the degenerate scene representation of Fig.~\ref{fig:teaser} (left) is unlikely to correspond to real scene geometry as the left half of the room is dispersed into space, and the estimated camera trajectory penetrates a wall.
Following this intuition, we reformulate the training objective of SCR to easily incorporate reconstruction priors in a maximum likelihood framework.
These priors can be hand-crafted \cite{Niemeyer2021Regnerf, roessle2022depthpriorsnerf, barron2022mipnerf360, xiong2023sparsegs} or  learned from data \cite{wynn2023diffusionerf}, and should lead to coherent and plausible reconstructions as the one in  Fig.~\ref{fig:teaser} (right).

For hand-crafted priors, we model plausible distributions of depth values and force reconstructed scene coordinates to follow those distributions.
For a learned prior, we pre-train a 3D diffusion model on point clouds of indoor scenes, to encode plausible scene geometries.
Training generative models, like diffusion models, on 3D representations is a difficult endeavor.
3D datasets are orders of magnitude smaller than their 2D counterparts.
The largest 3D datasets are comprised of simplistic 3D assets \cite{shapenet2015,objaverse,objaverseXL} wheres datasets with entire 3D scenes are smaller still, containing a few hundred scenes at most \cite{dai2017scannet, yeshwanthliu2023scannetpp, arnold2022mapfree, MegaDepthLi18}.
We also still lack efficient but expressive architectures to generate entire 3D scenes with high fidelity.
Both data and architecture limitations contribute to the fact that, so far, 3D point cloud diffusion was only shown for individual, isolated objects \cite{luo2021diffusion,Zhou2021PVD,Lyu2022diffusionrefinement,melaskyriazi2023pc2}. 
However, we show that for mere use as a \emph{prior}, a lean 3D point cloud diffusion model trained on relatively little data (around 700 indoor scenes) suffices to provide useful guidance when reconstructing indoor scenes. 

\noindent We summarize our \textbf{contributions}:
\begin{itemize}
    \item We reformulate SCR training as maximum likelihood learning to enable incorporation of reconstruction priors.
    \item We propose hand-crafted priors that push depth values of reconstructed scene coordinates to follow a reasonable distribution, and a learned prior in form of a 3D point cloud diffusion model that pushes reconstructed scene coordinates towards representing plausible scene layouts. 
    \item We plug our priors into state-of-the-art SCR frameworks ACE \cite{brachmann2023ace}, GLACE \cite{wang2024glace} and ACE0 \cite{brachmann2024acezero}.
They lead to learning better scene representations signified by more coherent and accurate point clouds, higher registration rates in SfM, and better pose estimates. 
Our approach does not significantly increase the training time of SCR models and it does not affect efficiency at query time. 
    \item As a byproduct, we show how measured depth maps can be incorporated as a prior, leading to effective RGB-D versions of ACE and ACE0.
\end{itemize}

\section{Related Work}

\paragraph{Visual Reconstruction and Relocalization}

Structure-from-motion refers to the problem of estimating scene geometry and camera poses from a set of images \cite{brown2005unsupervised, Snavely2006photo, Wu2013visualSfM, schonberger2016structure, pan2024glomap}.
Visual relocalization is a related task where the camera pose of a query image should be estimated relative to mapping images with known camera poses \cite{Sattler11, Sattler2017AS, sarlin2019coarse, kapture2020}.
Both tasks have been addressed with classical feature-matching where scenes are represented as sparse 3D point clouds.
The current generation of learning-based feature matchers \cite{sarlin20superglue, sun2021loftr, edstedt2024roma, lindenberger2023lightglue} is very reliable \cite{Jin2020imc}, with methods like MicKey \cite{barroso2024mickey} and MASt3R \cite{wang2024dust3r, leroy2024mast3r} demonstrating successful feature matching even for opposing shots.
However, despite the robustness of feature matchers, classical triangulation relies on the availability of sufficient multi-view constraints to reconstruct a scene \cite{hartley2003multiple}.
In texture-poor areas or with small camera baselines, point triangulation might still degenerate.
Since feature-based pipelines consist of various stages with complex control flow \cite{schonberger2016structure, Wu2013visualSfM, Snavely2006photo, wang2024vggsfm, he2024dfsfm}, it is difficult to infuse high-level priors for regularization.

Feed-forward reconstruction methods, like DUSt3R \cite{wang2024dust3r} and MASt3R \cite{leroy2024mast3r}, do suffer less from potential degeneracies of point triangulation.
Trained on large datasets, these methods incorporate strong reconstruction priors to enable reconstruction in sparse-view scenarios with little to no visual overlap. 
However, these methods are inherently binocular, and are difficult to scale to large image collections.
In a spirit similar to DUSt3R/MASt3R, our work aims at enabling strong priors for SCR methods.

As another alternative to feature matching, pose regression methods aim to predict camera poses using a feed-forward network \cite{kendall2015posenet,Kendall17,Brahmbhatt18,Shavit21multiscene, chen21}.
These networks represent the scene implicitly, and do not offer introspection on whether their scene representations are more or less prone to degeneracies.
In any case, pose regression methods either lack accuracy \cite{sattler2019limits}, or depend on massive amounts of scene-specific synthesized training data \cite{Moreau21lens, chen2022dfnet}.

\paragraph{Scene Coordinate Regression (SCR)}

SCR models fuse the key point extraction and matching stages of classical feature-based methods into a single regression step, performed by a scene-specific machine learning model.
Originally proposed by Shotton \etal \cite{shotton2013scene} for relocalization in small-scale indoor scenes with \mbox{RGB-D} images and random forests,
the approach was later extended to on-the-fly adaptation \cite{cavallari2017fly,Cavallari2019cascade,Cavallari2019network}, RGB inputs \cite{brachmann2016, Brachmann2018dsacpp}, using neural networks \cite{brachmann2017dsac, Brachmann2021dsacstar, brachmann2019ngransac} and to larger scenes \cite{Brachmann2019ESAC, li2020hscnet, Wang2024hscnetpp, wang2024glace}.
Recently, the ACE framework \cite{brachmann2023ace} has demonstrated training of extremely memory-efficient SCR models in a few minutes per scene.
Subsequently, ACE0 \cite{brachmann2024acezero} extended the applicability of ACE models from visual relocalization to structure-from-motion by demonstrating self-supervised training.

Most previous SCR methods rely entirely on scene-specific training, and make very little use of scene-agnostic data to distill reconstruction priors.
ACE and ACE0 use a pre-trained feature encoder.
However, as a low-level component, the feature encoder can do little to ensure coherence of the final scene representation. 
Marepo \cite{chen2024marepo} pre-trains a scene-agnostic transformer to replace RANSAC \cite{fischler1981random} and PnP \cite{gao2003complete} in the ACE pipeline.
In a similar spirit, SACReg \cite{Revaud2024sacreg} and SANet \cite{Luwei2019sanet} pre-train scene-agnostic SCR predictors that learn to interpolate scene coordinate annotations of mapping images. 
Marepo, SACReg and SANet all concern test-time components of SCR, and assume that a valid scene representation has already been built.
To the best of our knowledge, we are the first to regularize SCR training by high-level priors, derived from scene-agnostic data.
Some of our priors, regularizing the depth distribution of scene coordinates, bear conceptional similarity with depth regularization terms proposed for novel view synthesis \cite{roessle2022depthpriorsnerf, barron2022mipnerf360}.
We show how to apply such priors in the context of SCR, based on a probabilistic re-formulation of training.

\paragraph{Denoising Diffusion Models (DDMs)}

One of the priors we propose is a DDM on 3D point clouds to assess the likelihood of scene coordinate configurations.
DDMs are generative models that incrementally transform random noise to a target distribution, \eg transforming high-dimensional Gaussian noise to the distribution of natural images \cite{SohlDickstein2015diffusion, Ho2020DDPM, song2021ddim}.
The transformation is implemented using a model that is trained to remove a small amount of noise at each timestep.
Generation can be unconditional, or conditional, \eg generating images of a pre-defined class \cite{Dhariwal2021guidance, ho2021classifierfree}.
Adjacent to our work, scene coordinates have been used as a conditioning signal for image diffusion \cite{nguyen2025pointmapdiff}.

Due to the limited size and diversity of 3D datasets, diffusion for 3D generation is difficult.
2D diffusion models have been utilized to regularize or guide 3D generation \cite{poole2023dreamfusion, wu2024reconfusion, gao2024cat3d}.
For autonomous driving, large-scale scene generation has been demonstrated for 2.5D LiDAR range images \cite{zyrianov2022learning, ran2024towards, nunes2024lidarcompletion}, and coarse voxel grids \cite{Lee2023DiffusionPM, liu2024pyramiddiffusionfine3d}.  

Diffusion models are related to score matching methods \cite{yang2019datagradients, song2021scorebased} where models estimate the gradient of the log-likelihood of the target distribution \cite{Ho2020DDPM, Luo2022UnderstandingDM, Ruojin2020ShapeGF}.
This observation was used by DiffusioNeRF \cite{wynn2023diffusionerf} to utilize a diffusion model to represent the prior probability over RGB-D image patches when training a neural radiance field \cite{mildenhall2020nerf}.
We take inspiration from DiffusioNeRF's formulation but avoid the significant slow-down of rendering 2.5D image patches throughout training, and regularize directly in 3D.

\paragraph{DDMs for Point Clouds}
Luo and Hu \cite{luo2021diffusion} first demonstrated diffusion on 3D point clouds using a PointNet \cite{qi2016pointnet} architecture.
Subsequent works \cite{Zhou2021PVD, Lyu2022diffusionrefinement,zeng2022lion,mo2023ditd} proposed various improvements, including better architectures like PointNet++ \cite{qi2017pointnetplusplus} and point-voxel CNN \cite{liu2019pvcnn}.
These works model simple, isolated objects like \eg those of ShapeNet \cite{shapenet2015}.
Melas-Kyriazi \etal \cite{melaskyriazi2023pc2} show image-conditional diffusion of realistic, but still isolated, objects on Co3D \cite{reizenstein21co3d}.
To the best of our knowledge, generation of scene-level 3D point clouds using diffusion has not been demonstrated.
We use a point-voxel CNN \cite{liu2019pvcnn} to model the distribution of scenes in ScanNet \cite{dai2017scannet}.
While our generative model also cannot generate scenes with high fidelity, we show that it still serves as an efficient prior for regularizing SCR.

\begin{figure*}
    \centering
    \includegraphics[width=1.0\linewidth]{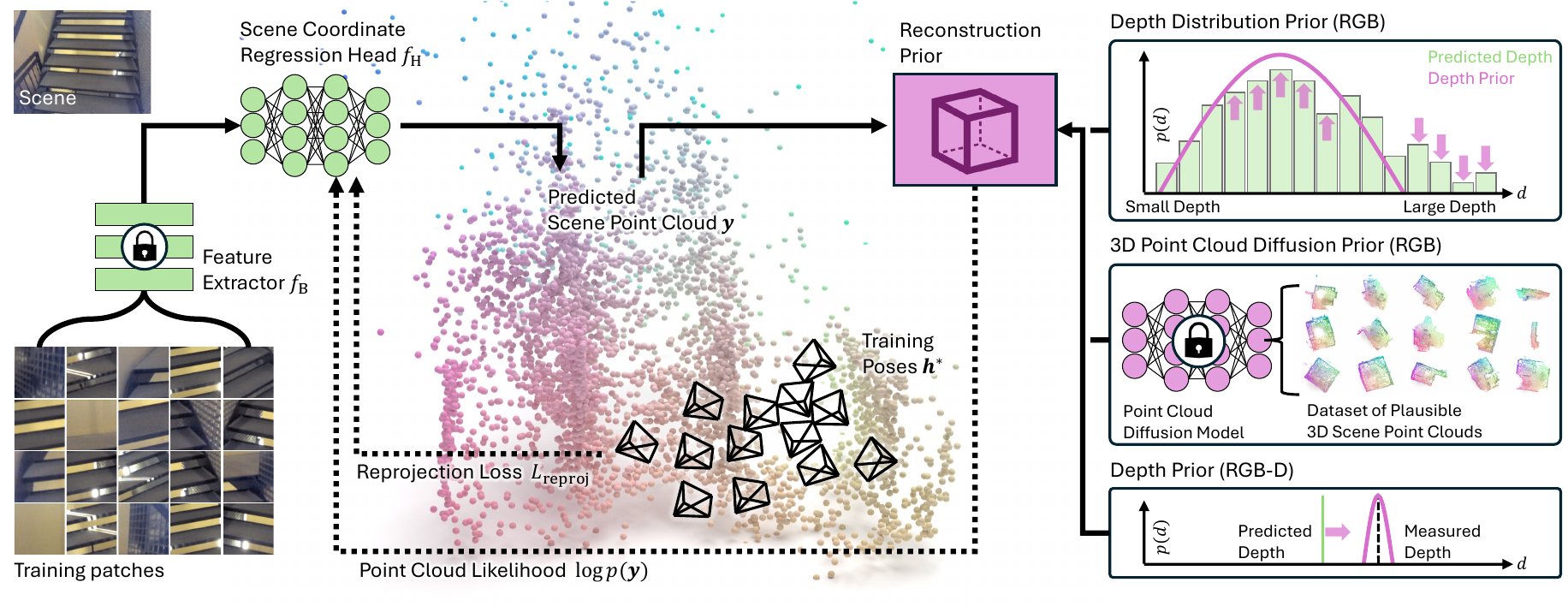}
    \vspace{-0.5cm}
    \caption{\textbf{System. Left:} We train a scene coordinate regression (SCR) model to represent a scene. Following ACE \cite{brachmann2023ace} and ACE0 \cite{brachmann2024acezero}, the SCR network predicts scene coordinates for a batch of random image patches in each training iteration. We supervise the SCR network with the reprojection error of scene coordinates \wrt ground truth camera poses. \textbf{Right:} Extending \cite{brachmann2023ace, brachmann2024acezero}, we introduce a reconstruction prior that outputs the gradient of the log-likelihood of the predicted point cloud. The prior guides the SCR model to learn a scene representation that is more likely to correspond to a real scene. We investigate various priors: A \emph{depth distribution prior} encourages depth values of reconstructed scene coordinates to follow a plausible distribution. A \emph{3D point cloud diffusion prior} is a network that was trained offline on a large corpus of scenes and encodes plausible scene layouts. We can also infuse a \emph{depth prior} if inputs are RGB-D.}
    \label{fig:framework}
\end{figure*}

\section{Method}
SCR models~\cite{shotton2013scene} encode a scene into a scene-specific neural network $f$. 
The network $f$ maps an image patch $\mathbf{p}_i$ centered around pixel $i$ of image $\mathcal{I}$ to a 3D scene point $\mathbf{y}_i$,
\begin{equation}
    \mathbf{y}_i = f(\mathbf{p}_i; \mathbf{w}),
\end{equation}
where $\mathbf{w}$ denotes the parameters that encode the scene.

An explicit 3D point cloud of the scene can be extracted from $f$ by running it over scene images and collecting the 3D points $\mathbf{y}_i$ \cite{Brachmann2021dsacstar}.
However, the main application of SCR models is camera pose estimation, where $f$ is applied to query images with unknown poses.
Since the prediction $\mathbf{y}_i$ represents a 2D-3D correspondence from pixel $i$ to scene space, the outputs of $f$ can be used for camera pose estimation by feeding them into RANSAC \cite{fischler1981random} and PnP \cite{gao2003complete}.

An SCR model $f$ can be trained using RGB-D or RGB images \cite{Brachmann2021dsacstar}. 
Although we will present an RGB-D version of our approach in Sec.~\ref{sec:prior:rgbd}, we are mainly interested in the general case and will assume that only RGB images are available if not stated otherwise.
Training also requires known camera poses for the training images.
With multiple mapping images $\mathcal{I}_\text{M}$ and their known camera poses, $f$ can be optimized using a reprojection loss $L_\text{reproj}$.
However, when multi-view constraints from the images are insufficient or ambiguous, $f$ may (partly) degenerate, and estimate scene points $\mathbf{y}_i$ that are noisy, distorted or plain outliers, and degrade downstream performance, see Fig.~\ref{fig:teaser} (left).

To mitigate this, we complement $L_\text{reproj}$ with a regularization term $L_\text{reg}$. 
$L_\text{reg}$ should enforce prior geometric constraints, \eg that scene coordinates follow a plausible depth distribution, or that all scene coordinates represent a coherent scene layout.
To this end, we reformulate the SCR training objective as maximum likelihood learning,
and set the regularization term to the negative log-likelihood of scene coordinates: $L_\text{reg} = -\log p(\mathbf{y})$. 
Since we apply the prior during training, it does not affect test time efficiency.
We base our work on ACE~\cite{brachmann2023ace}, due to its good accuracy and attractive training time. 
Our priors can be readily applied to any derivative of ACE as long as it keeps the general training framework.
We show results using ACE0 \cite{brachmann2024acezero}, a self-supervised version of ACE for SfM, and GLACE \cite{wang2024glace} for relocalization.
See Fig.~\ref{fig:framework} for a system overview.

\subsection{ACE Framework}
\label{sec: ace}
ACE decomposes a SCR model $f$ into a scene-agnostic feature extractor $f_\text{B}$ and a scene-specific regression head $f_\text{H}$,
\begin{equation}
    f(\mathbf{p}_i;\mathbf{w}) = f_\text{H}(\mathbf{f}_i; \mathbf{w}_\text{H}), \; \text{with}\, \mathbf{f}_i=f_\text{B}(\textbf{p}_i; \textbf{w}_\text{B}).
\end{equation}
The feature extractor $f_\text{B}$ maps an image patch to a high-dimensional feature vector $\mathbf{f}_i$, and the regression head $f_\text{H}$ maps the feature to a scene coordinate $\mathbf{y}_i$.
Following ACE, we assume $f_\text{B}$ was pre-trained and remains fixed. 
We focus on the mapping stage which optimizes $f_\text{H}$, see Fig.~\ref{fig:framework} (left).

ACE first passes all mapping images $\mathcal{I}_\text{M}$ to the feature extractor $f_\text{B}$ to obtain a training buffer with features for a large number of patches randomly sampled from mapping images. 
When training $f_\text{H}$, to achieve faster and more stable convergence, the buffered features are shuffled at each epoch.
In each training iteration, a batch of $N$ features $\mathbf{f} = \{\mathbf{f}_i\mid i=1, ..., N\}$ is sampled from 
the feature buffer to estimate the 3D scene points $\mathbf{y} = \{\mathbf{y}_i\mid i=1, ..., N\}$ for the corresponding 2D pixels from $\mathcal{I}_\text{M}$.

The ACE training loss $L_\text{reproj} (\mathcal{I}_\text{M}, \mathbf{y}, \mathbf{h}^*)$ is a pixel-wise projection loss applied to the 3D scene predictions $\mathbf{y}$ in each training iteration. 
Here, $\mathbf{h}^* = \{\mathbf{h}^*_i\mid i=1, ..., N\}$ denotes the ground truth poses for each image from which the pixels are sampled.
To regularize training in the first few iterations, ACE applies an additional loss, $L_\text{init} (\mathcal{I}_\text{M}, \mathbf{y}, \mathbf{h}^*)$, that mainly ensures that predicted coordinates are in front of their associated camera.
For further details, please refer to ACE~\cite{brachmann2023ace}.

\subsection{Probabilistic Training Objective}
\label{sec: prior_reg}

We recast the usual loss minimization objective of SCR as the maximization of a scene coordinate's probability.
The posterior probability for the scene points $\mathbf{y}$, given the mapping images $\mathcal{I}_\text{M}$ and poses $\mathbf{h}^*$, is proportional to the product of the likelihood $p(\mathbf{h}^*, \mathcal{I}_\text{M} \mid \mathbf{y})$ and the prior $p(\mathbf{y})$:
\begin{equation}
    p(\mathbf{y} \mid \mathbf{h}^*, \mathcal{I}_\text{M}) = \frac{p(\mathbf{h}^*, \mathcal{I}_\text{M} \mid \mathbf{y}) p(\mathbf{y})} {p(\mathbf{h}^*, \mathcal{I}_\text{M})}.
\end{equation}
Taking the negative logarithm of the posterior and omitting the constant $p(\mathbf{h}^*, \mathcal{I}_\text{M})$ yields:
\begin{equation}
\label{eq: posterior1}
    -\log p(\mathbf{y} \mid \mathbf{h}^*, \mathcal{I}_\text{M}) \propto -\log p(\mathbf{h}^*, \mathcal{I}_\text{M} \mid \mathbf{y}) - \log p(\mathbf{y}).
\end{equation}
In ACE mapping, minimizing the reprojection error can be interpreted as maximizing the log-likelihood of the mapping views and poses given the predicted scene points, \ie 
\begin{equation}
\label{eq: likelihood}
    -\log p(\mathbf{h}^*, \mathcal{I}_\text{M} \mid \mathbf{y}) \coloneq L_\text{reproj}.
\end{equation}
Similarly, ACE's regularization term can be interpreted as a form of prior on scene coordinates, \ie $- \log p(\mathbf{y}) \coloneq L_\text{init}$. 
However, ACE only optimizes either $L_\text{reproj}$ or $L_\text{init}$ per scene coordinate where the latter serves as an initialization target for the former.
In contrast, our perspective suggests that reprojection error and prior should be optimized jointly to maximize the likelihood of a scene coordinate.
We propose a set of alternative priors as stronger regularization terms  $L_\text{reg}$ instead of $L_\text{init}$, to be jointly optimized with the reprojection error: $L_\text{reproj} + L_\text{reg} = L_\text{reproj} - \log p(\mathbf{y})$.

\subsection{Depth Distribution Prior (RGB)}
\label{sec:prior:depth:distribution}

Firstly, we present a prior that builds on, and extends, the intuition of ACE's original initialization loss.
That loss ensures that the depth $d_i$ of a predicted scene coordinate $\mathbf{y}_i$ is within sensible bounds, namely $0 < d_i < d_\text{max}$, where $d_\text{max}$ is a user-defined upper bound.
Between those bounds, the initialization loss provides no signal. 
Instead ACE switches to optimizing the reprojection error only.

We propose to model a continuous distribution of plausible depth values, and to encourage scene coordinate to adhere to that distribution.
First, we have to choose a distribution family.
We need our prior to be defined over the whole real line to be able to even assess the likelihood of \emph{negative} depth values which can occur during SCR training.
We empirically found a Laplacian distribution, $\text{Lap}(d \mid \mu, b)$, to be a good choice. 
Having chosen the distribution family, we fit its hyper-parameters, mean $\mu$ and bandwidth $b$, to a held out set of scenes with measured (\ie ground truth) depth values.

We can directly use the negative log-likelihood of individual depth values as a prior per pixel $i$, \ie
\begin{equation}
\label{eq:prior:nll:rgb}
\log p(\mathbf{y_i}) \coloneq \lambda_\text{reg}~\log \text{Lap}(d_i \mid \mu, b),
\end{equation}
where $\lambda_\text{reg}$ is a hyper-parameter to balance the prior with the reprojection loss.
This prior corresponds to an $L_1$ loss on the predicted scene coordinate's depth, pulling it to the empirical mean.

While Eq.~\ref{eq:prior:nll:rgb} increases the likelihood of \emph{individual} scene coordinates, it does not guarantee the set of \emph{all} scene coordinates to actually follow the target distribution, in particular to have the correct variance.
Therefore, as an alternative to Eq.~\ref{eq:prior:nll:rgb}, we can utilize a \emph{distribution loss} between the predicted depth values and the prior.
We use the Wasserstein distance between the set of predicted depth values of a mini-batch, $\{d_i\}$, and the target Laplace distribution:
\begin{equation}
\log p(\mathbf{y_i}) \coloneq \lambda_\text{reg}~W\left[\{d_i\},~\text{Lap}(\cdot \mid \mu, b)\right].
\end{equation}
Here, we take advantage of the ACE framework which constructs mini-batches associated with random scene points throughout optimization.
Therefore, we can assume that the distribution of predicted depth values of a mini-batch roughly follows the depth distribution of the entire scene.

\subsection{Depth Prior (RGB-D)}
\label{sec:prior:rgbd}

The log-likelihood prior of Eq.~\ref{eq:prior:nll:rgb} lends itself to an extension that ingests measured depth values $d_i^*$ if available, namely for RGB-D input images. 
In this case, we substitute the broad distribution over plausible depth values with a narrow distribution centered at the measured depth value for each pixel:
\begin{equation}
\label{eq:prior:nll:rgbd}
\log p(\mathbf{y_i}) \coloneq \lambda_\text{reg}~\log \text{Lap}(d_i \mid d_i^*, b').
\end{equation}
Here, the bandwidth $b'$ is a user-defined value that controls the tolerance of the prior. We will use $b'= 10$cm.

\subsection{Point Cloud Diffusion Prior (RGB)}
\label{sec: diff_prior}

Finally, we propose adding a 3D diffusion prior to SCR training.
The prior is pre-trained on a held out set of scenes, and encodes knowledge about plausible scene point clouds.
During SCR training, the diffusion model is kept fixed, and nudges scene coordinates towards coherent scene layouts.

In forward diffusion, Gaussian noise $\epsilon$ is progressively added to a signal $\mathbf{x}_0$ over $T$ time steps until the signal becomes completely noisy. 
Similarly, SCR training incrementally recovers the scene point cloud from a random initialization.
Therefore, SCR training aligns with reversed diffusion, and a 3D diffusion model can guide SCR optimization.

Diffusion models~\cite{song2021ddim, Ho2020DDPM} learn to recover the original signal $\mathbf{x}_0$ by estimating the noise added to the noisy signal $\mathbf{x}_\tau$ at time step $\tau \in [0, T-1]$, through a neural network $\epsilon_\theta$. 
It has been shown in \cite{wynn2023diffusionerf, Ho2020DDPM, vincent2011connection} that the noise estimate is proportional to the score function of the input signal, \ie
\begin{equation}
\label{eq: prior}
     \nabla_\mathbf{x}\log p(\mathbf{x}) \coloneq - \lambda_\text{reg} \epsilon_\theta(\mathbf{x}_\tau, \tau).
\end{equation}
In our context, this allows us to utilize a diffusion denoising model $\epsilon_\theta$ as prior over 3D coordinates $\mathbf{y}$, because the prediction of the model corresponds to the gradient of the log-likelihood of scene coordinates.
This prior is inspired by DiffusioNeRF \cite{wynn2023diffusionerf} with the main difference that we regularize directly in 3D. 
\begin{table*}[]
\scriptsize
\centering
\caption{\textbf{Reconstruction of ScanNet \cite{dai2017scannet} and Indoor6 \cite{do2022learning}.} We compare results of ACE0 without and with our priors added. We report the percentage of images registered to the reconstruction (\emph{Reg. Rate}), absolute trajectory and relative pose errors (\emph{ATE}/\emph{RPE}) as well as median pose errors compared to pseudo ground truth (pGT), and PSNR of novel view synthesis using two separate evaluation splits: the usual split with one test frame / seven training frames, and a harder split alternating between 60 test frames and 60 training frames.  }
\label{tab:exp:sfm}
\begin{tabular}{@{}llccccccccc@{}}
\cmidrule[1pt](r){1-8} \cmidrule[1pt](l){10-11}
                                &                                     &                      & \multicolumn{2}{c}{Comparison to Pose pGT (BundleFusion)} &  & \multicolumn{2}{c}{Splatfacto PSNR (dB) $\uparrow$} &  & \multicolumn{2}{c}{Indoor6}      \\ \cmidrule(lr){4-5} \cmidrule(lr){7-8} \cmidrule(l){10-11} 
                                &                                     & Reg. Rate $\uparrow$ & ATE / RPE (cm) $\downarrow$      & Med. Err. (cm/°) $\downarrow$      &  & 1/7                 & 60/60               &  & Reg. Rate $\uparrow$ & PSNR (dB) $\uparrow$ \\ \cmidrule(r){1-8} \cmidrule(l){10-11} 
\multirow{5}{*}{RGB}   & ACE0                                & 98.1\%               & 26.6 / 4.0                       & 19.7 / 9.0                         &  & 30.2                & 22.3                &  & 57.1\%               & 13.5                 \\
                                & \textbf{ACE0 + Laplace NLL} (Ours)  & \textbf{98.9\%}      & 25.4 / \textbf{3.5}              & 17.5 / 8.8                         &  & 30.2                & 22.2                &  & 58.0\%               & 14.1                 \\
                                & \textbf{ACE0 + Laplace WD} (Ours)   & 98.7\%               & 25.9 / 3.6                       & 17.5 / 6.8                         &  & 30.3                & 21.7                &  & 57.7\%               & 14.1                 \\
                                & \textbf{ACE0 + Diffusion} (Ours)    & 98.6\%               & 26.5 / 3.8                       & 18.8 / 8.9                         &  & 30.2                & 22.4                &  & \textbf{61.8\%}      & \textbf{14.6}        \\ \cmidrule(r){1-8} \cmidrule(l){10-11} 
\multirow{2}{*}{RGB-D} & ACE0 + DSAC* Loss                   & 96.2\%               & 29.2 / 6.0                       & 20.9 / 5.9                         &  & 30.0                & 21.9                &  & N/A                  & N/A                  \\
                                & \textbf{ACE0 + Laplace NLL} (Ours)  & \textbf{98.9\%}      & \textbf{18.3} / \textbf{3.5}     & \textbf{12.8 / 4.4}                &  & \textbf{30.6}       & \textbf{22.9}       &  & N/A                  & N/A                  \\ \cmidrule[1pt](r){1-8} \cmidrule[1pt](l){10-11} 
\end{tabular}
\end{table*}

\label{sec: diff_model}
\paragraph{Architecture} Our noise estimator $\epsilon_\theta(\mathbf{x}_\tau, \tau)$ takes the noisy point cloud $\mathbf{x}_\tau \in \mathbb{R}^{N \times 3}$ and the diffusion timestep $\tau$ as inputs.
We adopt PVCNN~\cite{liu2019point} with certain modifications made by ~\cite{melaskyriazi2023pc2} for diffusion timestep embedding.

\paragraph{Training} We follow the training protocol in DDPM~\cite{Ho2020DDPM}. 
During each forward iteration, we first normalize the input point cloud $\mathbf{x}_0$ with a predefined scaling factor to ensure the points are within the range $[-1, 1]$. We then transform $\mathbf{x}_0$ to a noisy version $\mathbf{x}_\tau$ by adding noise to $\mathbf{x}_0$ according to the noise schedule at a randomly sampled time step $\tau$.
The training objective of the diffusion model is expressed as:
\begin{equation}
\label{eq: diffusion2}
    L_\text{training} = \mathbb{E}_{\tau, \mathbf{\epsilon} \sim \mathcal{N}(0, 1), \mathbf{x}_0} \left[\|( \epsilon - \epsilon_\theta(\mathbf{x}_\tau, \tau))\|^2_2\right] .
\end{equation}

\paragraph{Inference}
Once the diffusion model is trained, we integrate it into SCR mapping.
We take the estimated scene points $\mathbf{y}$ as input for the diffusion model and estimate the noise.
We use the estimate as regularization as specified by \cref{eq: prior}.
The distributions of the Gaussian noise and the noise encountered during ACE mapping differ, particularly during the first mapping iterations. 
Therefore, we apply the diffusion prior only after iteration 5k of ACE mapping.
We align the diffusion time $\frac{T}{20}$ with ACE iteration 5k, and linearly interpolate timestep $\tau$ down to 0 as training progresses.
We do not apply the diffusion prior to points with a reprojection error smaller than 30 pixels, as we assume that sufficient multi-view constraints exist for those points.

\section{Experiments}

\paragraph{Implementation Details}
We build on the official PyTorch~\cite{paszke2017automatic} implementation of ACE0~\cite{brachmann2024acezero}.
Unless otherwise specified, we use the same parameters as ACE0 \cite{brachmann2024acezero} for reconstruction, and the same parameters as ACE \cite{brachmann2023ace} for relocalization. 
We use the ACE feature backbone which was pre-trained on 100 ScanNet training scenes.
We also integrate the diffusion prior into GLACE~\cite{wang2024glace}. 
In this variant, we maintain the regularization identical to its implementation in ACE, while keeping all other settings of GLACE.

\paragraph{Training the Priors}
We fit our priors to the training set of ScanNetV2~\cite{dai2017scannet} which consists of 706 scenes, densely reconstructed from RGB-D images.
For the depth distribution prior (Sec.~\ref{sec:prior:depth:distribution}), we fit a Laplace distribution to randomly sampled depth values of the training images, yielding a mean  of $\mu=1.73$m and a bandwidth of $b=60$cm.
For training the diffusion model (Sec.~\ref{sec: diff_prior}), we use the ground truth point cloud of each training scene from ScanNetV2 as target.
In each iteration, we randomly sample 5,120 points from a point cloud in the training dataset and apply augmentations including random rotation, translation and scaling within a certain range. 
The total number of diffusion time steps is set to 200.
We train the diffusion model with a batch size of 16 for 100,000 iterations on a single V100 GPU. 
We use AdamW~\cite{kingma2014adam}, with a learning rate that decays linearly from 0.0002 to 0 throughout the training process.
The diffusion model is frozen after training and applied during the mapping stage across all scenes and datasets.

\subsection{Structure-from-Motion}

We reconstruct a number of indoor scenes using ACE0 \cite{brachmann2024acezero}, with and without incorporating our priors.

\paragraph{Datasets}
We evaluate on the first 20 test scenes of ScanNetV2~\cite{dai2017scannet}, and on the mapping sequences of Indoor6~\cite{do2022learning}.
Where ScanNet consists of single room scans, Indoor6 features six larger indoor environments comprised of multiple rooms.
Indoor6 does not come with RGB-D images.

\paragraph{Metrics}
We report the percentage of images successfully registered to the reconstruction (\emph{Reg.~Rate}), \ie images with a final inlier count above 1,000 \cite{brachmann2024acezero}.
We confirm the quality of estimated camera poses using novel view synthesis \cite{waechter2017rephotography,brachmann2024acezero}.
That is, we estimate camera poses using all images.
Then, we divide images into training and validation images for Splatfacto \cite{nerfstudio,ye2024gsplat,kerbl3Dgaussians} to check whether we can re-render the scene based on the estimated camera poses.
We report \emph{PSNR} numbers on the usual training/validation split of Splatfacto, taking one validation image for every seven training images. 
For ScanNet we also report results for a harder split, alternating between 60 validation images and 60 training images.
Finally, for ScanNet, we compare the estimated poses to the pseudo ground truth (pGT, \cite{brachmann2021limits}) that comes with the dataset, estimated by BundleFusion \cite{dai2017bundlefusion}, a RGB-D SLAM system that exploits the ordering of images.
For reference, BundleFusion achieves a PSNR of 22.2dB on the 60/60 Splatfacto split, \emph{lower} that some of our results.
Still, SLAM poses serve as a suitable reference in terms of global consistency, and we report the absolute trajectory error (ATE) and relative pose errors (RPE) \cite{sturm12iros}, as well as median rotation and translation errors after least-squares alignment to the pGT camera trajectory.

\begin{table*}[t!]
\scriptsize
\centering
\caption{\textbf{Relocalization Results on 7Scenes~\cite{shotton2013scene}.} We report the percentage of test images below a 5cm/5$^\circ$ pose error (higher is better), mapping time and map size. Methods in ``SCR w/ 3D" use depth or 3D point cloud supervision during mapping. Best results within the SCR groups are highlighted in \textbf{bold}. All methods use SfM mapping poses.}
 \resizebox{\textwidth}{!}{
\begin{tabular}{clcccccccccc}
\toprule
Type                                                                         & Method                            & Mapping Time          & Map Size    & Chess   & Fire   & Heads     & Office  & Pumpkin & Redkitchen & Stairs & Avg    \\ \midrule
\multirow{3}{*}{\rotatebox{90}{FM}}                                                          & AS (SIFT)~\cite{sattler2016efficient}                         & $\sim$1.5h & $\sim$200MB &  N/A       & N/A       &  N/A         &    N/A     &       N/A  &       N/A     &   N/A     & \textbf{98.5\%} \\
& D.VLAD+R2D2~\cite{humenberger2020robust}                       & $\sim$1.5h                      & $\sim$1GB   &     N/A    & N/A       &    N/A       &   N/A      &   N/A      &    N/A        &    N/A    & 95.7\% \\
& hLoc (SP+SG)~\cite{sarlin2019coarse, sarlin20superglue}                      & $\sim$1.5h                      & $\sim$2GB   &  N/A       &    N/A    &    N/A       &   N/A      &     N/A    &   N/A         &    N/A    & 95.7\% \\ \midrule \midrule
\multirow{5}{*}{\rotatebox{90}{\begin{tabular}[c]{@{}c@{}}SCR w/ 3D\end{tabular}}} & DSAC*~\cite{Brachmann2021dsacstar}                             & 15h                   & 28MB        & 99.8\%  & 98.9\% & 99.8\%    & 98.5\%  & 98.9\%  & 97.8\%     & 93.8\% & \textbf{98.2\%} \\
& SRC~\cite{dong2022visual}                               & 2min                  & 40MB        & 89.1\%  & 79.6\% & 97.6\%    & 84.1\%  & 65.7\%  & 87.3\%     & 64.7\% & 81.1\% \\
& FocusTune~\cite{nguyen2024focustune}                         & 6min                  & 4MB         & 99.7\%  & 99.0\% & 87.1\%    & 99.9\%  & \textbf{99.9\%}  & \textbf{100\%}    & \textbf{99.5\%} & 97.9\% \\ 
& ACE \cite{brachmann2023ace, brachmann2024acezero} + DSAC* Loss \cite{Brachmann2021dsacstar} & 5.5min                  & 4MB         & \textbf{100\%} & \textbf{99.5\%} & \textbf{100\%} & 98.8\% & 96.2\% & 98.2\% & 82.3\% & 96.4\% \\ 
& \textbf{ACE + Laplace NLL} (Ours) & 5.5min                  & 4MB         & \textbf{100\%} & 99.2\% & \textbf{100\%} & \textbf{99.8\%} & 97.2\% & 98.4\% & 87.3\% & 97.4\% \\ \midrule 
\multirow{7}{*}{\rotatebox{90}{SCR}}                                                         & DSAC* ~\cite{Brachmann2021dsacstar}                            & 15h                   & 28MB        & \textbf{100\%} & 99.1\%          & 98.8\%           & 99.3\%           & 99.4\%           & 96.9\%          & 78.6\%          & 96.0\%          \\
& GLACE~\cite{wang2024glace}                            & 6min                  & 9MB         & \textbf{100\%} & \textbf{99.9\%} & 99.9\%           & 99.8\%           & \textbf{100\%} & 98.2\%          & 71.2\%          & 95.6\%          \\
& \textbf{GLACE + Diffusion} (Ours)                             & 9min                  & 9MB         & \textbf{100\%} & 99.8\% & \textbf{100\%}           & 99.9\%           & 99.5\% & 98.8\%          & 73.6\%          & 95.9\%          \\
& ACE~\cite{brachmann2023ace}                             & 5min                  & 4MB         & \textbf{100\%} & 99.5\%          & 99.7\%           & \textbf{100\%} & 99.9\%           & 98.6\%          & 81.9\%          & 97.1\%          \\
& \textbf{ACE + Laplace NLL} (Ours) & 4.5min                  & 4MB         & \textbf{100\%} & 98.9\% & 99.9\% & 99.9\% & \textbf{100\%} & 98.5\% & 84.2\% & 97.3\% \\
& \textbf{ACE + Laplace WD} (Ours) & 4.5min                  & 4MB         & \textbf{100\%} & 99.6\% & \textbf{100\%} & 99.9\% & 99.8\% & 98.2\% & 83.2\% & 97.2\% \\
& \textbf{ACE + Diffusion} (Ours) & 8min                  & 4MB         & \textbf{100\%} & 99.5\%          & \textbf{100\%} & \textbf{100\%} & 99.0\%           & \textbf{99.1\%} & \textbf{86.2\%} & \textbf{97.7\%} \\ \bottomrule
\end{tabular}
}
\label{table: 7scenes}
\end{table*}

\paragraph{Discussion}
We report results in Table \ref{tab:exp:sfm}.
We couple our Laplace depth distribution prior (Sec.~\ref{sec:prior:depth:distribution}) with log-likelihood optimization (\emph{Laplace NLL}) as well as the Wasserstein distance (\emph{Laplace WD}).
Both variations of the prior increase the registration rates and pose quality on ScanNet.
The novel view synthesis quality is largely on par with ACE0, except for a noticeable drop in PSNR for the 60/60 evaluation split when using the Wasserstein distance.
Effects are stronger on Indoor6 where ACE0 struggles due to the size of the scenes. 
Both depth distribution priors lead to higher registration rates, and increase PSNR by 0.6 dB.

The diffusion prior yields slight but consistent improvements across all metrics on ScanNet.
On Indoor6, the diffusion prior causes an increase in registered images (+4.7\%) leading to higher PSNR (+1.1 dB).
The diffusion prior takes the global scene layout into account, providing a stronger regularization signal than our depth distribution priors.

On ScanNet, we also test a version of ACE0 with our RGB-D prior (Sec.~\ref{sec:prior:rgbd}).
We compare to an ACE0 RGB-D baseline where we activate the ACE0 RGB-D loss, usually only applied when initializing the reconstruction, throughout the entire reconstruction.
The ACE0 RGB-D loss is inspired by DSAC* \cite{Brachmann2021dsacstar} and optimizes the distance of predicted scene coordinates to ground truth derived from depth maps.
The loss switches to the reprojection error, when predicted scene coordinates are within 10cm of the ground truth.
This RGB-D baseline of ACE0 (denoted \emph{ACE0 + DSAC* Loss}) performs worse than the default ACE0 on ScanNet, presumably due to the hard switch between RGB and RGB-D terms reducing optimization stability.
In contrast, ACE0 coupled with our depth prior, derived from our probabilistic interpretation of SCR optimization, yields significant improvements across all metrics.

\subsection{Relocalization Results}
\label{sec: reloc_results}

We evaluate performance of our priors on the relocalization task using the 7Scenes and Indoor6 datasets, comparing it against other Scene Coordinate Regression (SCR) methods and feature-matching (FM) approaches.
We report the percentage of images localized within a 5cm/5$^\circ$ threshold \cite{shotton2013scene}.

\begin{figure*}[t!]
    \centering
    \includegraphics[width=0.99\linewidth]{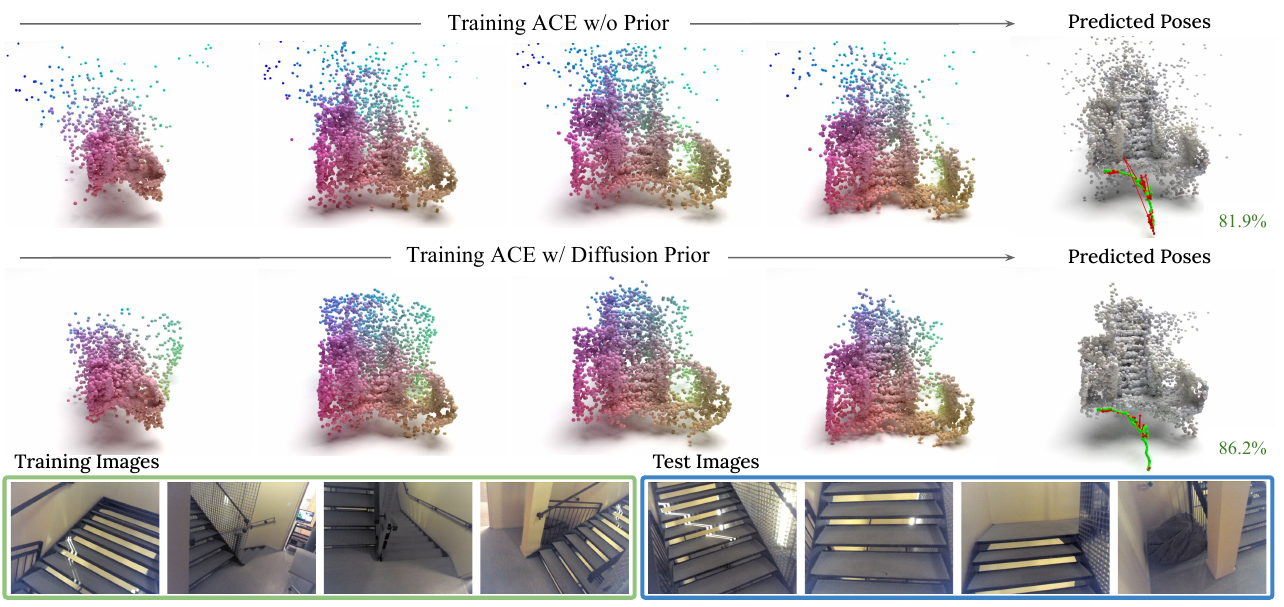}
    \caption{\textbf{Diffusion Prior on Stairs.} We guide the training of ACE \cite{brachmann2023ace} on the Stairs scene with our 3D diffusion model, leading to a more coherent scene geometry, and higher pose estimation accuracy on test images (5cm,5$^\circ$ threshold).}
    \label{fig:exp:stairs}
\end{figure*}

\paragraph{7Scenes} 
Results in \cref{table: 7scenes} demonstrate that ACE coupled with our priors achieves higher relocalization accuracy than ACE alone. 
The diffusion prior has the strongest effect, improving results by 4.1\% on the most difficult scene, \emph{Stairs}. 
The supplement includes results using alternative ground truth poses \cite{brachmann2021limits} where the same trend holds.
In Fig.~\ref{fig:exp:stairs}, we show qualitatively how the diffusion prior leads to a more compact representation for this scene.
To underline the versatility of our priors, we report results of GLACE \cite{wang2024glace}, a recent ACE extension, coupling it with our diffusion prior.
Again, relocalization results improve on average, particularly on the Stairs scene.
In the supplement, we analyze the effect of our priors on the point clouds learned by ACE.

In terms of mapping time, our depth distribution priors lead to simplified optimization objectives compared to ACE, resulting in a slight reduction in mapping time.
On the other hand, incorporating the diffusion prior incurs additional cost, as we need to execute a separate model many times throughout ACE mapping.
However, the nominal increase in mapping time is still modest, with +3 minutes.

\input{tables/indoor6}

\begin{figure*}[h!]
    \centering
    \includegraphics[width=1.0\linewidth]{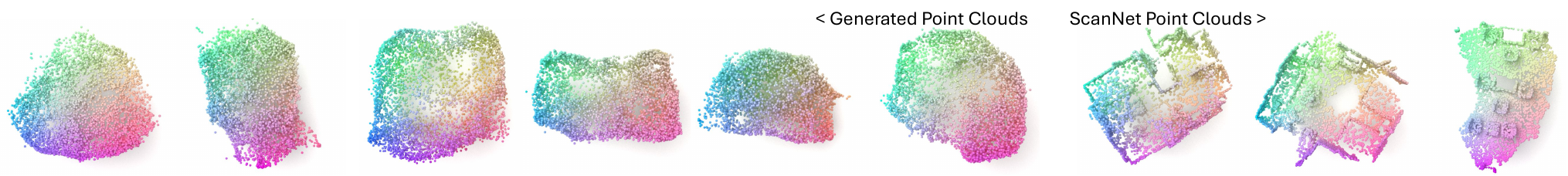}
    \caption{\textbf{Generated Point Clouds.}
    Point clouds generated by our diffusion model together with ScanNet point clouds used for training.
    }
    \label{fig:prior_samples}
\end{figure*}

\paragraph{Indoor6}
In \cref{table: cd}, we report the average relocalization accuracy on Indoor6 for ACE~\cite{brachmann2023ace}; GLACE~\cite{wang2024glace} with and without our diffusion prior. 
Our experiments show that increasing the batch size significantly improves the localization accuracy, presumably due to the larger size of the scenes.
Therefore, we evaluated two versions of each method, using batch sizes of 5,120 and 51,200.
We observed noticeable variance in relocalization accuracy on this dataset, and report mean results over 5 runs.  
Adding our prior helps on average, although ultimate conclusions are difficult due to the variance on this dataset.
For per-scene results, please see the supplement.
For EGFS~\cite{liu2025reprojection}, another ACE derivative, the code is not publicly available yet.
We report results with batch size 5,120 from their paper. %, without knowing the batch size they use.
EGFS could be coupled with our diffusion prior, as well.

\subsection{Prior Introspection}
\cref{fig:prior_samples} shows point clouds generated by the diffusion model, compared with ScanNet scenes, illustrating the prior learned by the model.
Although lacking fine details, the prior represents sensible room layouts sufficient to regularize SCR training. 
The supplement includes further analysis, regarding point cloud encoder architecture, efficiency, the prior's effect on scarce data and hyper-parameters.

\section{Conclusion}
We have presented a probabilistic reformulation of scene coordinate regression training that allows for the easy incorporation of reconstruction priors.
We presented multiple potential priors: priors regularizing the distribution of reconstructed depth values, a prior leveraging measured depth, as well as a high-level learned prior in the form of a 3D point cloud diffusion model.
We found that the regularization terms can be integrated into ACE-based frameworks, such as ACE, ACE0 or GLACE, yielding performance gains while not affecting the test-time latency.

\noindent\textbf{Limitations~}
Our experiments have focused on indoor scenes, with results on a few outdoor scenes in the supplement.
Properly extending our approach to outdoor scenes requires better models of depth distribution, and more diverse data for diffusion training to learn a comprehensive prior over larger areas.
A point cloud encoding network with better expressiveness could result in generations with higher fidelity, but it would need to remain efficient to be practical.
An additional conditional signal for the diffusion model can also be a promising direction for future research. 

%% file: tables/indoor6.tex
\begin{table}[]
\footnotesize
\caption{\textbf{Relocalization Results on Indoor6~\cite{do2022learning}.} We report relocalization accuracy (5cm, 5$^\circ$) and mapping time. Except for EGFS, we report all results as mean over 5 runs. Best results \textbf{bold} within each pair w/ and w/o our diffusion prior (denoted \emph{-Diff}).}
\centering
\begin{tabular}{clcc}
\midrule[1pt]
\noalign{\smallskip}
& & Reloc.~Acc. & Map.~Time \\
\noalign{\smallskip}
\hline \noalign{\smallskip} 
\multirow{6}{*}{N=5,120}  & EGFS~\cite{liu2025reprojection} & 56.1\% & 21 min\\ 
\cdashline{2-4}\noalign{\smallskip} 
& GLACE~\cite{wang2024glace} & 44.2\% $\pm$ 1.8\% & 11 min\\
& \textbf{GLACE-Diff} (Ours) & \textbf{46.4}\% $\pm$ 1.9\% & 15 min \\ 
\cdashline{2-4}\noalign{\smallskip} 
& ACE~\cite{brachmann2023ace} &  36.2\% $\pm$ 1.5\% & 5 min\\
& \textbf{ACE-Diff} (Ours) & \textbf{37.5}\%  $\pm$ 1.8\% & 8 min\\ 
\hline \noalign{\smallskip} 
\multirow{4}{*}{N=51,200} & GLACE~\cite{wang2024glace} & 69.5\% $\pm$ 1.4\% & 33 min\\
& \textbf{GLACE-Diff} (Ours) & \textbf{69.6\%} $\pm$ 2.0\% & 40 min\\
\cdashline{2-4}\noalign{\smallskip} 
& ACE~\cite{brachmann2023ace} & 57.2\% $\pm$ 1.6\% & 10 min\\
& \textbf{ACE-Diff} (Ours) & \textbf{57.9}\% $\pm$ 1.1\% & 13 min\\
\midrule[1pt]
\end{tabular}
\label{table: cd}
\end{table}

%% file: supplement.tex
\clearpage
\setcounter{page}{1}
\maketitlesupplementary
\blfootnote{* Work done during an internship at Niantic.} %remove for arXiv

\section{Implementation Details}

\subsection{Diffusion Training}
The point clouds in ScanNetV2~\cite{dai2017scannet} are confined to the positive octant of the coordinate system, with the xy-plane aligned to the floor.
To enhance the diversity of the training data, we first re-center the point clouds by shifting their xy coordinates so that their centers align with the origin. During each forward iteration, we randomly sample 5,120 points from one scene and apply data augmentation by randomly rotating the point cloud along each axis, adding a random translation sampled from a normal distribution with zero mean and unit variance, and applying a random scaling factor sampled from a uniform distribution in the range $[0.5, 1.5)$. Before passing a point cloud to the diffusion model, we re-scale it with a scale factor of $20$ to ensure that most points lie within the range $[-1, 1]$. Finally, diffusion noise is added to the point cloud following the standard DDPM schedule~\cite{Ho2020DDPM}.

\subsection{SCR Mapping with Diffusion} 
During SCR mapping, the point clouds output by the Scene Coordinate Regression network are re-scaled using the same scale factor (20) as during training, before being fed into the diffusion model for noise estimation. If the batch size exceeds 5,120, we randomly subsample the points down to 5,120 to compute the diffusion regularization. This step is taken to prevent excessive processing time during point cloud encoding when the point cloud size is large. 
To balance the magnitude of diffusion regularization against the reprojection loss, we adopt the gradient normalization approach from DiffusionNeRF~\cite{wynn2023diffusionerf}. Specifically, the gradient of the regularization term is normalized with respect to itself and then scaled by a weight. 
In all experiments, this weight is set to 1,000, with a warm-up phase spanning the first 1,000 iterations after diffusion regularization begins at iteration 5,000 of SCR mapping. During this warm-up, the weight increases linearly from 0 to 1,000.

\subsection{Prior Weights}
For the depth distribution priors, we utilize a weight of $\lambda_\text{reg}=0.1$. 
For the depth prior using RGB-D images, we use a weight of $\lambda_\text{reg}=1$.
These weights have been found by monitoring the magnitude of gradients stemming from the priors and the reprojection error during some mapping runs on ScanNet training sequences.
For the weighting schema of the diffusion prior, see the previous section.

\subsection{Diffusion-ACE Alignment}
\begin{figure*}
    \centering
    \includegraphics[width=1.0\linewidth]{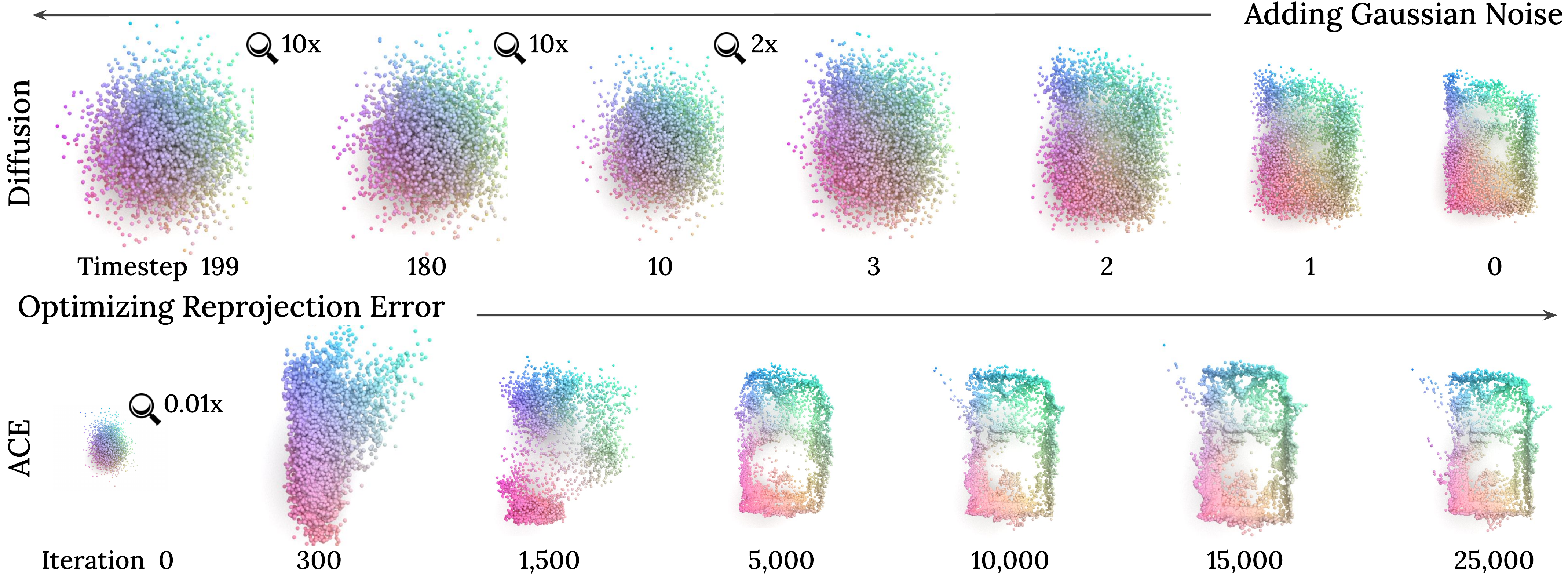}
    \caption{\textbf{Diffusion Process vs.~ACE Training.} The evolution of point clouds during ACE training does not match the forward diffusion process over the full range. Hence, we align the diffusion time steps 10-0 with the ACE iterations 5,000-25,000.}
    \label{fig:diffusion_mismatch}
\end{figure*}

See Fig.~\ref{fig:diffusion_mismatch} for a visualization of the ACE mapping process versus a reverse diffusion process on the same scene. 
The beginning of ACE mapping does not align well with diffusion, hence we apply the diffusion prior only after a stand-by time of 5k iterations.

\subsection{Point Cloud Visualization}
Our point cloud visualizations were generated using the visualization code from PointFlow~\cite{Yang2019pointflow}.

\subsection{Details of Baseline Approaches}
For ACE~\cite{brachmann2023ace} and GLACE~\cite{wang2024glace}, we use their public code to reproduce their results on 7Scenes which have slight differences with the results reported in their original papers.
Similarly, we obtain the results of ACE and GLACE on Indoor6 by running their code using the default settings.

\begin{table*}[h]
\scriptsize
\centering
\caption{\textbf{Relocalization Results on 7Scenes with SLAM Poses.} We report the percentage of test images below a 5cm/5$^\circ$ pose error, mapping time and map size. Methods in ``SCR w/ 3D" use depth or 3D point cloud supervision during mapping. Best results within the SCR groups are highlighted in \textbf{bold}.}
 \resizebox{\textwidth}{!}{
\begin{tabular}{clcccccccccc}
\toprule
Type                                                                         & Method                            & Mapping Time          & Map Size    & Chess   & Fire   & Heads     & Office  & Pumpkin & Redkitchen & Stairs & Avg    \\ \midrule
\multirow{4}{*}{\rotatebox{90}{FM}}                                                          & AS (SIFT)~\cite{sattler2016efficient}                         & $\sim$1.5h & $\sim$200MB &  N/A       & N/A       &  N/A         &    N/A     &       N/A  &       N/A     &   N/A     & 68.7\% \\
& D.VLAD+R2D2~\cite{humenberger2020robust}                       & $\sim$1.5h                      & $\sim$1GB   &     N/A    & N/A       &    N/A       &   N/A      &   N/A      &    N/A        &    N/A    & 77.6\% \\
& hLoc (SP+SG)~\cite{sarlin2019coarse, sarlin20superglue}                      & $\sim$1.5h                      & $\sim$2GB   &  N/A       &    N/A    &    N/A       &   N/A      &     N/A    &   N/A         &    N/A    & 76.8\% \\ 
& pixLoc~\cite{sarlin2021back}                      & $\sim$1.5h                      & $\sim$1GB   &  N/A       &    N/A    &    N/A       &   N/A      &     N/A    &   N/A         &    N/A    & 75.7\% \\ 
\midrule \midrule
\multirow{5}{*}{\rotatebox{90}{\begin{tabular}[c]{@{}c@{}}SCR w/ 3D\end{tabular}}} & DSAC*~\cite{Brachmann2021dsacstar}                             & 15h                   & 28MB        & \textbf{97.3}\% & 94.0\% & 99.7\% & \textbf{87.4\%} & \textbf{62.9\%}  & \textbf{63.7\%}     & \textbf{83.4\%} & \textbf{84.0\%} \\
& SANet~\cite{Luwei2019sanet}                         & 2.3min                  & 550MB    &  N/A       &    N/A    &    N/A       &   N/A      &     N/A    &   N/A         &    N/A    & 68.2\%\\
& SRC~\cite{dong2022visual}                               & 2min                  & 40MB        &  N/A       &    N/A    &    N/A       &   N/A      &     N/A    &   N/A         &    N/A    & 55.2\%\\
 & ACE \cite{brachmann2023ace, brachmann2024acezero} + DSAC* Loss \cite{Brachmann2021dsacstar} & 5.5min                  & 4MB         & 96.0\% & 94.0\% & \textbf{99.9\%} & 84.5\% & 55.1\% & 57.8\% & 76.6\% & 80.6\% \\ 
& \textbf{ACE + Laplace NLL} (Ours) & 5.5min                  & 4MB         & 97.1\% & \textbf{94.8\%} & 99.8\% & 86.4\% & 57.5\% & 59.2\% & 82.9\% & 82.5\% \\
 \midrule 
\multirow{7}{*}{\rotatebox{90}{SCR}}                                                         & DSAC* ~\cite{Brachmann2021dsacstar}                            & 15h                   & 28MB        & 95.3\% & \textbf{94.5\%} & 98.1\% & 86.3\% & 61.6\%  & 64.0\%     & 67.6\% & 81.1\%            \\
& GLACE~\cite{wang2024glace}                            & 6min                  & 9MB         & 98.5\%         & 93.7\% & 99.7\%         & \textbf{90.2\%}         & 61.9\%  & \textbf{73.9\%}          & 54.0\%          & 81.7\%            \\
& \textbf{GLACE + Diffusion} (Ours)                             & 9min                  & 9MB         & \textbf{98.9\%}         & 92.8\% & 99.3\%         & 89.5\%         & \textbf{64.3\%}  & 72.6\%          & 56.4\%          & \textbf{82.0\%}           \\
& ACE~\cite{brachmann2023ace}                             & 5min                  & 4MB         & 96.7\%         & 92.4\%          & 99.7\%         & 86.0\%         & 59.2\%         & 60.3\%          & 68.9\%          & 80.5\%          \\
& \textbf{ACE + Laplace NLL} (Ours) & 4.5min                  & 4MB         & 96.6\% & 92.8\% & 99.8\% & 85.0\% & 57.8\% & 59.2\% & 72.4\% & 80.5\% \\
& \textbf{ACE + Laplace WD} (Ours) & 4.5min                  & 4MB         & 96.7\% & 94.0\% & 99.6\% & 85.7\% & 59.7\% & 59.4\% & 71.0\% & 80.9\% \\
& \textbf{ACE + Diffusion} (Ours) & 8min                  & 4MB         
& 96.9\%         & 93.8\% & \textbf{99.8\%} & 85.7\% & 57.1\%  & 59.8\%     & \textbf{74.3\%} & 81.1\%  \\ \bottomrule
\end{tabular}
}
\label{table: 7scenes_slam}
\end{table*}

\begin{table*}[]
\scriptsize
\centering
\caption{\textbf{Per-Scene Relocalization Accuracy on Indoor6.} We report the percentage of test images below a 5cm, 5$^\circ$ pose error, mapping time and map size. Methods in “SCR w/ 3D” use the 3D point cloud as supervision during mapping. `50K' denotes a batch size of 51200 points and `dual'  represents an ensemble model with two clusters. Methods using our diffusion prior are denoted as \emph{-Diff}.}
\label{table: indoor6_reloc}
\begin{tabular}{clccccccccc}
\toprule
Type                                                                   & \multicolumn{1}{c}{Method} & Mapping Time & Map Size    & scene1 & scene2a & scene3 & scene4a & scene5 & scene6 & Average \\ \midrule
FM                                                                     & hLoc (SP+SG)               & $\sim$3.3    & $\sim$1.5GB & 70.5\% & 52.1\%  & 86.0\% & 75.3\%  & 58.0\% & 86.7\% & 71.4\%  \\ \midrule
\multirow{3}{*}{\begin{tabular}[c]{@{}c@{}}SCR\\ (w/ 3D)\end{tabular}} & DSAC*                      & 15h          & 28MB        & 18.7\% & 28.0\%  & 19.7\% & 60.8\%  & 10.6\% & 44.3\% & 30.4\%  \\
& SLD (300 LM)               & 5.5h         & 15MB        & 47.2\% & 48.2\%  & 56.2\% & 67.7\%  & 33.7\% & 52.0\% & 50.8\%  \\
& SLD (1000 LM)              & 44h          & 120MB       & 68.5\% & 62.6\%  & 76.2\% & 77.2\%  & 57.8\% & 78.0\% & 70.1\%  \\ \midrule
\multirow{15}{*}{SCR}                                                  & DSAC*                      & 15h          & 28MB        & 23.0\% & 33.9\%  & 26.0\% & 67.1\%  & 10.6\% & 50.2\% & 35.1\%  \\
& EGFS                       & 21min        & 4.5MB       & 46.4\% & 60.6\%  & 56.4\% & 78.7\%  & 22.8\% & 71.6\% & 56.1\%  \\
& EGFS (dual)                  & 21min        & 9MB         & 58.5\% & 59.1\%  & 67.0\% & 76.1\%  & 30.6\% & 75.9\% & 61.2\%  \\
\cdashline{2-11}\noalign{\smallskip} 
& GLACE                      & 11min        & 9MB         & 31.1\% & 44.8\%  & 37.3\% & \textbf{72.2\%}  & 19.4\% & 60.1\% & 44.2\% $\pm$ 1.8\%   \\
& \textbf{GLACE-Diff}        & 15min        & 9MB         & \textbf{35.7\%} & \textbf{46.5\%}  & \textbf{41.5\%} &  69.0\%  & \textbf{22.8\%} & \textbf{62.7\%} & \textbf{46.4\%} $\pm$ 1.9\% \\
\cdashline{2-11}\noalign{\smallskip} 
& ACE                        & 5min         & 4MB         & 24.5\% & 35.1\%  & 34.4\% & \textbf{58.9\%}  & 15.7\% & 48.4\% & 36.2\% $\pm$ 1.5\%  \\
& \textbf{ACE-Diff}          & 8min         & 4MB         & \textbf{26.9\%} & \textbf{35.3\%}  & \textbf{37.2\%} & 58.5\%  & \textbf{16.6\%} & \textbf{50.3\%} & \textbf{37.5\%} $\pm$ 1.8\% \\
\cdashline{2-11}\noalign{\smallskip} 
& GLACE (50K)                & 33min        & 9MB         & 64.1\% & \textbf{68.5\%}  & \textbf{73.1\%} & \textbf{84.8\%}  & 41.2\% & 85.2\% & 69.5\% $\pm$ 1.4\% \\
& \textbf{GLACE-Diff} (50K)  & 40min        & 9MB         & \textbf{65.1\%} & 67.4\%  & 73.0\% & 84.2\%  & \textbf{41.8\%} & \textbf{85.9\%} & \textbf{69.6\%} $\pm$ 2.0\% \\
\cdashline{2-11}\noalign{\smallskip} 
& ACE (50K)                  & 10min        & 4MB         & \textbf{47.9\%} & 55.0\%  & 60.5\% & \textbf{77.3\%}  & 26.4\% & 75.9\% & 57.2\%  $\pm$ 1.6\%\\
& \textbf{ACE-Diff}  (50K)   & 13min        & 4MB         & \textbf{47.9\%} & \textbf{56.0\%}  & \textbf{62.0\%} & 76.5\%  & \textbf{27.1\%} & \textbf{77.7\%} & \textbf{57.9\%}  $\pm$ 1.1\%\\ \cdashline{2-11}\noalign{\smallskip} 

& GLACE (dual)                  & 22min        & 18MB         & 51.1\% & 56.8\%  & 60.9\% & 75.9\%  & \textbf{30.5\%} & 75.6\% & 58.5\%  $\pm$ 1.9\%\\
& \textbf{GLACE-Diff}  (dual)   & 30min        & 18MB         & \textbf{51.6\%} & \textbf{58.2\%} & \textbf{61.9\%} & \textbf{76.7\%}  & 28.2\% & \textbf{76.5\%} & \textbf{58.9\%}  $\pm$ 1.7\%\\ 
\cdashline{2-11}\noalign{\smallskip} 
 & ACE (dual)                  & 10min        & 8MB   &     41.4\% & 46.0\% & 54.4\%  & 65.8\%  & 20.5\% & 63.1\% & 48.5\%  $\pm$ 1.9\%\\
& \textbf{ACE-Diff}  (dual)   & 16min        & 8MB         & \textbf{43.2\%} & \textbf{48.7\%}  & \textbf{55.0\%} & \textbf{66.1\%}  & \textbf{21.8\%} & \textbf{64.4\%} & \textbf{49.9\%}  $\pm$ 1.8\%\\ 
\bottomrule
\end{tabular}
\label{table: indoor6_per_scene}
\end{table*}

\begin{table*}[t]
\scriptsize
\caption{\textbf{Depth Evaluation on 7Scenes~\cite{shotton2013scene}.} We report the errors between the estimated depth images, derived from the 3D scene points generated by SCR, against the ground truth depth images for each method.}
\centering
\resizebox{0.7\textwidth}{!}{
\begin{tabular}{lccccccc}
\hline
\noalign{\smallskip}
                          & Abs Rel $\downarrow$ & Sq Rel $\downarrow$ & RMSE $\downarrow$ & RMSE log $\downarrow$ & $\delta_1 \uparrow$ & $\delta_2 \uparrow$ & $\delta_3 \uparrow$ \\ \hline

                          ACE~\cite{brachmann2023ace}   & 0.48                 & 24.25               & 2.26              & 0.33                  & 0.87                & 0.93                & 0.96                \\
                          \textbf{ACE+Laplace NLL} (Ours)  & 0.34        & 1.34       & 1.65     & 0.33                  & \textbf{0.88}                & 0.94       & \textbf{0.97}       \\ 
                          \textbf{ACE+Laplace WD} (Ours)  & 0.35        & 1.43       & 1.66     & \textbf{0.32}                  & \textbf{0.88}                & 0.94       & \textbf{0.97}       \\ 
                          \textbf{ACE+Diffusion} (Ours)  & \textbf{0.32}        & \textbf{1.17}       & \textbf{1.62}     & \textbf{0.32}                  & \textbf{0.88}                & \textbf{0.95}       & \textbf{0.97}       \\  \cdashline{1-8}\noalign{\smallskip} 
 GLACE~\cite{wang2024glace} & 0.40                 & 25.57               & 2.36              & 0.25         & 0.89       & 0.95       & 0.97     \\
 \textbf{GLACE+Diffusion} (Ours) & \textbf{0.28}                 & \textbf{0.96}               & \textbf{1.60}              & \textbf{0.23}         & \textbf{0.90}       & \textbf{0.96}       & \textbf{0.98}       \\
                          \hline
\end{tabular}
\label{table: depth}
}
\end{table*}

\begin{table*}[]
\scriptsize
\centering
\caption{\textbf{Ablation Study of the Diffusion Model Architecture.} We compare the relocalization accuracy (5cm, 5$^\circ$) and depth errors on 7Scenes with different architectures 
of the diffusion point cloud encoder.
} 
 \resizebox{0.45\textwidth}{!}{
\begin{tabular}{lcccc}
\toprule
              & Reloc Acc $\uparrow$ & RMSE $\downarrow$ &RMSE log $\downarrow$ & $\delta_1 \uparrow$  \\ \midrule
PVCNN~\cite{liu2019pvcnn}         &    \textbf{97.7\%} &  \textbf{1.62}&  \textbf{0.32} &\textbf{0.883} \\
Pointwise-Net~\cite{luo2021diffusion} &  97.5\% &  1.64& 0.69 &0.879 \\
 \bottomrule
\end{tabular}
\label{table: training_ablation}
}
\end{table*}

\begin{figure}[htb]
    \centering
    \includegraphics[width=0.95\linewidth]{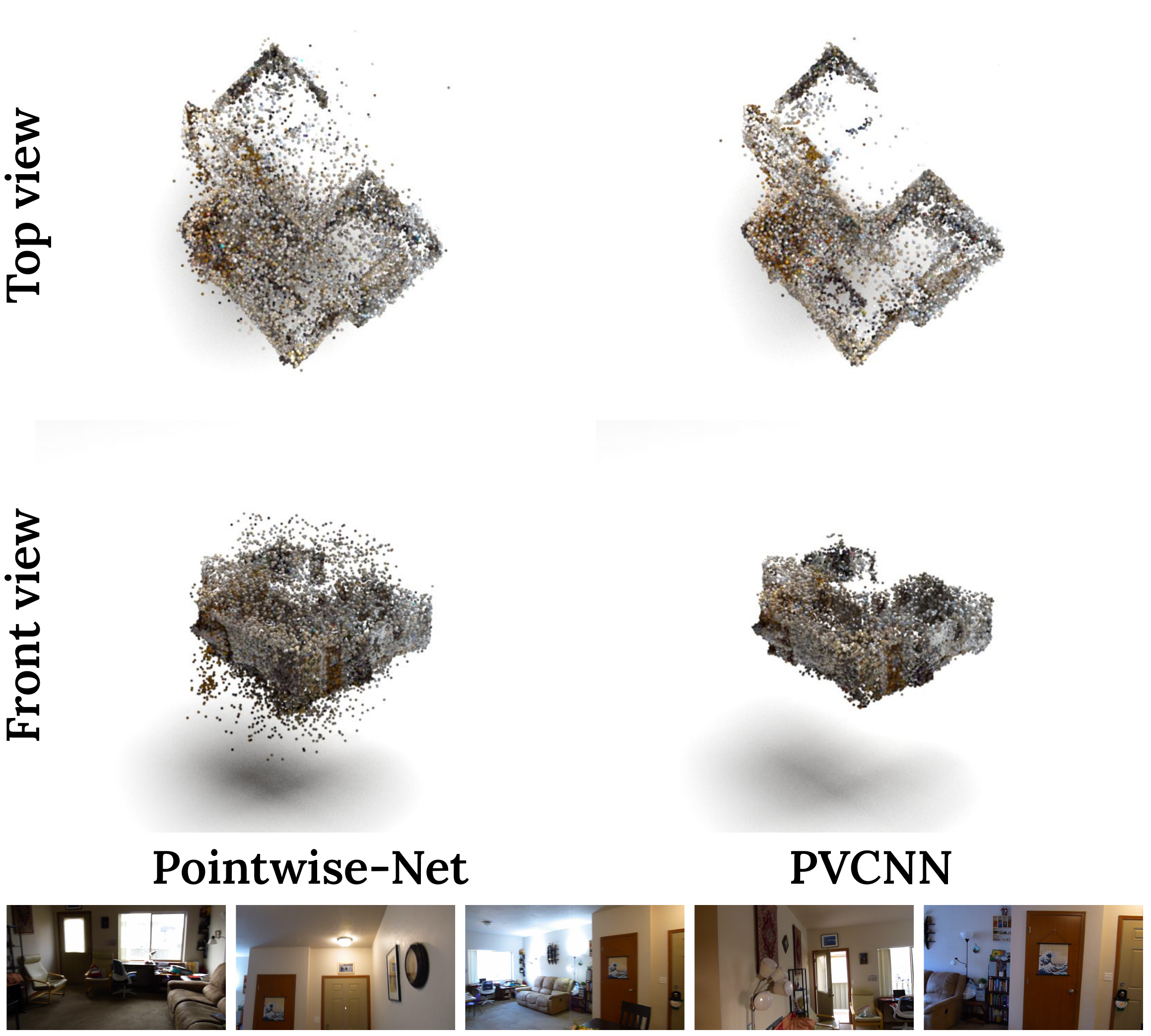}
    \caption{\textbf{Qualitative Results with Different Architectures for Point Cloud Encoder.} We compare the point clouds obtained after SCR mapping using diffusion regularization with different point cloud encoders. Pointwise-Net encodes every point independently whereas PCVNN captures structural information.
    }
    \label{fig:encoder_ablation}
\end{figure}

\begin{table*}[]
\scriptsize
\centering
\caption{\textbf{Mask Threshold for Diffusion.} We compare the relocalization accuracy (5cm, 5$^\circ$) on 7Scenes with different mask thresholds $\kappa$ for applying diffusion regularization.}
\begin{tabular}{lcccccccc}
\toprule
   & Chess            & Fire            & Heads            & Office           & Pumpkin         & Redkitchen      & Stairs          & Avg             \\ \midrule
ACE  & 100.0\% & 99.5\%          & 99.7\% & 100.0\%          & 99.9\% & 98.6\%          & 81.9\%          & 97.1\%          \\
\midrule
$\kappa = 0$  & \textbf{100.0\%} & 99.4\%          & \textbf{100.0\%} & 99.8\%           & \textbf{99.3\%} & 98.4\%          & 85.4\%          & 97.5\%          \\
$\kappa = 30$ & \textbf{100.0\%} & \textbf{99.5\%} & \textbf{100.0\%} & \textbf{100.0\%} & 99.0\%          & \textbf{99.1\%} & \textbf{86.2\%} & \textbf{97.7\%} \\
$\kappa = 60$ & \textbf{100.0\%} & 99.6\%          & 99.9\% & 99.8\%           & 99.0\%          & 98.3\%          & 80.8\%          & 96.8\%          \\ 

\bottomrule
\end{tabular}
\label{table: mask_threshold}
\end{table*}

\begin{figure}
    \centering
    \includegraphics[width=1.0\columnwidth]{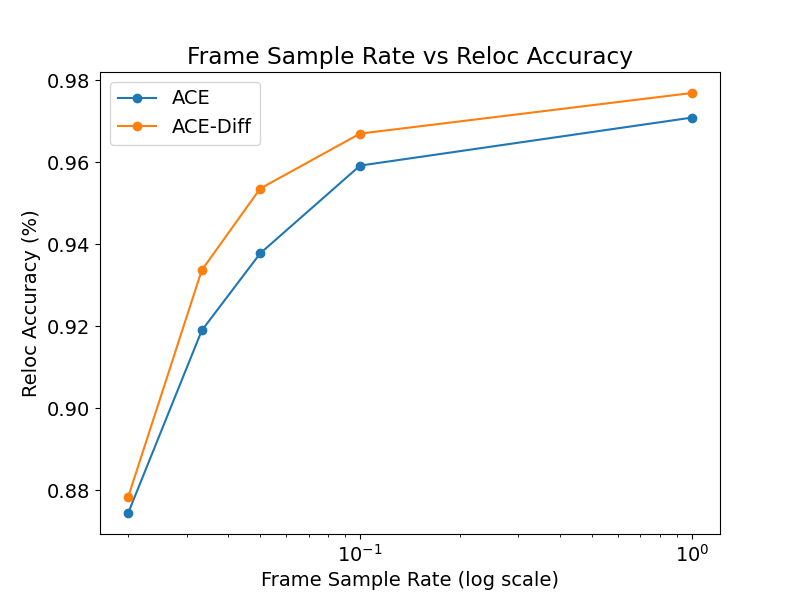}
    \caption{\textbf{Mapping Sample Rate vs.~Accuracy on 7Scenes.} Subsampling the mapping frames leads to a decrease in relocalization performance, but ACE with our diffusion prior (\emph{ACE-Diff}) consistently outperforms ACE.
    }%
    \label{fig: sparse}
\end{figure}

\section{Additional Results}

\subsection{Relocalization Results}

\paragraph{7Scenes}
We conduct experiments with an alternative set of ground truth poses \cite{brachmann2021limits}  stemming from an RGB-D SLAM system \cite{shotton2013scene}.
The relocalization performance is shown in \cref{table: 7scenes_slam}.
Analogously with the results obtained using SfM poses, presented in the main paper, our priors improve performance over the baseline models (ACE and GLACE) especially on the Stairs scene.

\paragraph{Indoor6}
We report the per-scene relocalization results in \cref{table: indoor6_per_scene} for ACE and GLACE with and without our diffusion prior. 
Results are averaged over 5 runs, except for EGFS~\cite{liu2025reprojection}, where results are taken from its paper. 
Additionally, we include model variants where `50K' denotes a batch size of 51,200 points and `dual'  represents an ensemble of two models trained on a pre-clustering of the mapping camera poses \cite{brachmann2023ace}.
For each variant, our model outperforms its respective baseline, demonstrating improved relocalization accuracy. 
Among SCR approaches, GLACE with our diffusion prior achieves the highest accuracy on this dataset.

\subsection{Depth Evaluation on 7Scenes}

\label{sec: recon_results}
We assess the point cloud quality after ACE mapping with and without our priors.
To this end we compare depth values derived from the predicted scene coordinates with the ground truth depth images.
In line with previous works~\cite{liu2015learning, fu2018deep, ranftl2021vision}, we apply standard metrics, including: Abs Rel, Sq Rel, RMSE, RMSE log, $\delta_1$, $\delta_2$ and $\delta_3$.
They are defined as follows:
\vspace{0.3cm}
\begin{itemize}
    \item Abs Rel: $\frac{1}{|\mathcal{V}|} \sum_{d \in \mathcal{V}} \|d - d_\text{gt}\| / d_\text{gt}$;
    \item Sq Rel: $\frac{1}{|\mathcal{V}|} \sum_{d \in \mathcal{V}} \|d - d_\text{gt}\|^2_2 / d_\text{gt}$;
    \item RMSE: $\sqrt{\frac{1}{|\mathcal{V}|} \sum_{d \in \mathcal{V}} \|{d} -{d_\text{gt}}\|^2_2}$;
    \item RMSE log: $\sqrt{\frac{1}{|\mathcal{V}|} \sum_{d \in \mathcal{V}} \|\log{d} - \log{d_\text{gt}}\|^2_2}$;
    \item $\delta_i$: $\%$ of $\mathcal{V}$ s.t. $ \text{max}(\frac{d}{d_\text{gt}}, \frac{d_\text{gt}}{d}) = \delta < i$;
\end{itemize}
\vspace{0.3cm}
where $d$ is the estimated depth, $d_\text{gt}$ is the ground truth depth, and $\mathcal{V}$ is the collection of all valid pixels on a depth map.

To generate per-frame depth maps we compute the scene coordinates for each mapping view, transform them into camera space, and derive the estimated depth from the z-coordinates.
The average errors between these depth estimates and the ground truth across all scenes in 7Scenes are presented in \cref{table: depth}. 
The results demonstrate that all our priors significantly reduce the depth error compared to the baseline approaches for both ACE and GLACE.
In particular, the number of outlier points decreases as signified in the drastic reduction of the Sq Rel metric.

\subsection{Analysis}

\paragraph{Point Cloud Encoder Architecture}
We compare the PVCNN architecture~\cite{liu2019point} with the Pointwise-Net used in~\cite{luo2021diffusion}. Pointwise-Net processes each point independently, limiting its ability to capture structural relationships. As shown in \cref{table: training_ablation} and \cref{fig:encoder_ablation}, PVCNN outperforms Pointwise-Net in both relocalization and reconstruction, showing the advantage when incorporating structural information.

\paragraph{Mask Threshold}
As described in the main paper, during SCR mapping, diffusion regularization is applied only to points with a reprojection error greater than 30 pixels. \cref{table: mask_threshold} compares the relocalization accuracy when regularization is applied to all points ($\kappa = 0$) and only to points with reprojection errors above 60 pixels ($\kappa = 60$). In the latter scenario, regularization affects fewer points, resulting in performance that closely resembles the original ACE.
While applying diffusion to all points also improves upon the baseline ACE, the results indicate that $\kappa = 30$ achieves the best accuracy.

\paragraph{Frame Sample Rate}
\cref{fig: sparse} 
shows the impact of sub-sampling the number of mapping images in 7Scenes. Our diffusion prior mitigates the effect of scarce data.

\begin{table}
\scriptsize
\centering
\caption{\textbf{Efficiency-Accuracy Analysis for Regularization}. We analyze how the frequency of applying diffusion during mapping impacts both relocalization accuracy (5cm, 5$^\circ$) and mapping time.}
\resizebox{0.45\textwidth}{!}{
\begin{tabular}{ccccccc}
\hline \noalign{\smallskip}
\multirow{2}{*}{$k$} &  & \multicolumn{2}{c}{7Scenes} & \multirow{2}{*}{} & \multicolumn{2}{c}{Indoor6 (N=51200)} \\ \cline{3-4} \cline{6-7} \noalign{\smallskip}
                   &  & Reloc Acc $\uparrow$      & Time       &                   & Reloc Acc $\uparrow$         & Time             \\ \hline
1                  &  & 97.7\%         & 18min      &                   & 57.5\%             & 24min            \\
2                &  & 97.7\%         & 11min      &                   & 57.5\%             & 17min            \\
4                &  & 97.7\%         & 8min       &                   & 57.9\%             & 13min          \\
8                &  & 97.6\%         & 6min       &                   & 57.2\%             & 12min          \\ \hline
\end{tabular}
\label{table: efficency}
}
\end{table}

\paragraph{Efficiency}
The diffusion regularization adds extra computation time to each ACE mapping iteration.
To balance efficiency and accuracy, we reduce the frequency of applying the diffusion regularization by only implementing it every $k$ mapping iterations. 
As shown in \cref{table: efficency}, setting $k=4$ (i.e. running one diffusion iteration every 4 mapping steps) achieves the optimal trade-off, adding approximately 3 minutes to the total ACE mapping time.

\subsection{Outdoor Scenes}
Our priors were designed for indoor scenes.
Outdoor scenes pose significant additional challenges.
For example, the distribution of depth for outdoor scenes can be more complex, multi-modal and vary tremendously from scene to scene.
Generative modeling of outdoor scenes requires more expressive architectures that are in turn computationally more demanding, and would slow down ACE mapping.
Outdoor environments come with a significant level of diversity, and require large datasets to learn priors that generalize.

Still, we show some promising results on outdoor scenes using our \emph{indoor} priors.
We evaluate ACE with and without our priors on the Cambridge Landmarks dataset~\cite{kendall2015posenet}, which consists of outdoor scenes of varying extent spanning a small shop facade to an entire university court. 
As shown in \cref{tab:outdoor}, our priors can lead to small improvements. 
For the depth distribution prior, we use a Laplace distribution with with a mean of 25m and a bandwidth of 10m.
For the diffusion prior, we employ the same model used in our other experiments, which is trained on \emph{indoor} ScanNet data.
For the RGB-D prior, we set a bandwidth of 0.5m, and use MVS depth maps for mapping published by~\cite{Brachmann2018dsacpp}.

\begin{table}[h!]
\resizebox{\columnwidth}{!}{
    \begin{tabular}{@{}clcccccc@{}}
    \toprule
    \multicolumn{1}{l}{} &                          & GC            & KC   & OH   & SF           & StMC & Mean  \\ \midrule
    \multirow{4}{*}{RGB} & ACE                      & 41.7          & 27.0 & 30.0 & 5.4          & 20.4 & 24.9 \\
                         & ACE + Laplace NLL & \textbf{39.6} & 28.3 & 28.0 & 5.3          & 21.6 & 24.5 \\
                         & ACE + Laplace WD  & 50.1          & 26.2 & 32.1 & 6.0          & 20.1 & 26.9 \\
                         & ACE + Indoor Diff.   & 45.6          & 27.9 & 28.5 & \textbf{5.2} & 20.8 & 25.6 \\ \hdashline
    RGB-D & ACE + Laplace NLL & 49.3 & \textbf{22.9} & \textbf{24.2} & 5.7 & \textbf{15.4} & \textbf{23.5} \\ \bottomrule
    \end{tabular}
}
\caption{\textbf{Outdoor Scenes}. Median position error (cm) on Cambridge Landmarks. Scene names abbreviated by first letters.}
\label{tab:outdoor}
\end{table}

\subsection{More Qualitative Result}
\begin{figure}[]
  \centering
  \includegraphics[width=0.99\linewidth]{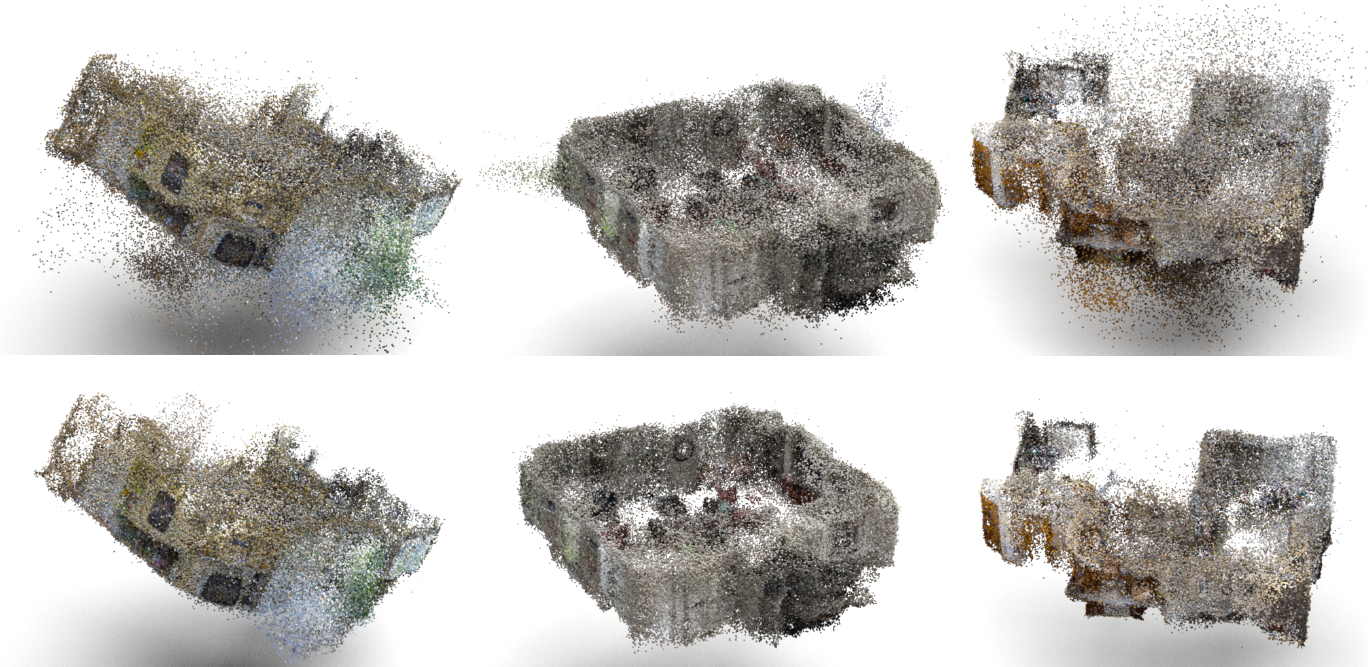}
   \caption{\textbf{Qualitative Results on Indoor6.} Point clouds extracted from the ACE (top) and ACE+Diffusion (bottom) networks.}
   \label{fig:indoor6}
\end{figure}

We present a qualitative comparison between ACE and ACE enhanced with the diffusion prior in \cref{fig:indoor6}, where the effectiveness of the diffusion prior is evident, particularly in reducing noise.

%% file: main.bbl
\begin{thebibliography}{111}
\providecommand{\natexlab}[1]{#1}
\providecommand{\url}[1]{\texttt{#1}}
\expandafter\ifx\csname urlstyle\endcsname\relax
  \providecommand{\doi}[1]{doi: #1}\else
  \providecommand{\doi}{doi: \begingroup \urlstyle{rm}\Url}\fi

\bibitem[Arnold et~al.(2022)Arnold, Wynn, Vicente, Garcia-Hernando, Monszpart, Prisacariu, Turmukhambetov, and Brachmann]{arnold2022mapfree}
Eduardo Arnold, Jamie Wynn, Sara Vicente, Guillermo Garcia-Hernando, {\'{A}}ron Monszpart, Victor~Adrian Prisacariu, Daniyar Turmukhambetov, and Eric Brachmann.
\newblock Map-free visual relocalization: Metric pose relative to a single image.
\newblock In \emph{ECCV}, 2022.

\bibitem[Barron et~al.(2022)Barron, Mildenhall, Verbin, Srinivasan, and Hedman]{barron2022mipnerf360}
Jonathan~T. Barron, Ben Mildenhall, Dor Verbin, Pratul~P. Srinivasan, and Peter Hedman.
\newblock {Mip-NeRF 360}: Unbounded anti-aliased neural radiance fields.
\newblock In \emph{CVPR}, 2022.

\bibitem[Barroso-Laguna et~al.(2024)Barroso-Laguna, Munukutla, Prisacariu, and Brachmann]{barroso2024mickey}
Axel Barroso-Laguna, Sowmya Munukutla, Victor Prisacariu, and Eric Brachmann.
\newblock Matching {2D} images in {3D}: Metric relative pose from metric correspondences.
\newblock In \emph{CVPR}, 2024.

\bibitem[Brachmann and Rother(2018)]{Brachmann2018dsacpp}
Eric Brachmann and Carsten Rother.
\newblock Learning less is more - 6d camera localization via {3D} surface regression.
\newblock In \emph{CVPR}, 2018.

\bibitem[Brachmann and Rother(2019{\natexlab{a}})]{Brachmann2019ESAC}
Eric Brachmann and Carsten Rother.
\newblock Expert sample consensus applied to camera re-localization.
\newblock In \emph{ICCV}, 2019{\natexlab{a}}.

\bibitem[Brachmann and Rother(2019{\natexlab{b}})]{brachmann2019ngransac}
Eric Brachmann and Carsten Rother.
\newblock Neural-guided {RANSAC}: Learning where to sample model hypotheses.
\newblock In \emph{ICCV}, 2019{\natexlab{b}}.

\bibitem[Brachmann and Rother(2021)]{Brachmann2021dsacstar}
Eric Brachmann and Carsten Rother.
\newblock Visual camera re-localization from {RGB} and {RGB-D} images using {DSAC}.
\newblock \emph{IEEE TPAMI}, 2021.

\bibitem[Brachmann et~al.(2016)Brachmann, Michel, Krull, Yang, Gumhold, and Rother]{brachmann2016}
Eric Brachmann, Frank Michel, Alexander Krull, Michael~Y. Yang, Stefan Gumhold, and Carsten Rother.
\newblock Uncertainty-driven 6{D} pose estimation of objects and scenes from a single {RGB} image.
\newblock In \emph{CVPR}, 2016.

\bibitem[Brachmann et~al.(2017)Brachmann, Krull, Nowozin, Shotton, Michel, Gumhold, and Rother]{brachmann2017dsac}
Eric Brachmann, Alexander Krull, Sebastian Nowozin, Jamie Shotton, Frank Michel, Stefan Gumhold, and Carsten Rother.
\newblock {DSAC}-differentiable {RANSAC} for camera localization.
\newblock In \emph{CVPR}, 2017.

\bibitem[Brachmann et~al.(2021)Brachmann, Humenberger, Rother, and Sattler]{brachmann2021limits}
Eric Brachmann, Martin Humenberger, Carsten Rother, and Torsten Sattler.
\newblock On the limits of pseudo ground truth in visual camera re-localisation.
\newblock In \emph{ICCV}, 2021.

\bibitem[Brachmann et~al.(2023)Brachmann, Cavallari, and Prisacariu]{brachmann2023ace}
Eric Brachmann, Tommaso Cavallari, and Victor~Adrian Prisacariu.
\newblock Accelerated coordinate encoding: Learning to relocalize in minutes using {RGB} and poses.
\newblock In \emph{CVPR}, 2023.

\bibitem[Brachmann et~al.(2024)Brachmann, Wynn, Chen, Cavallari, Monszpart, Turmukhambetov, and Prisacariu]{brachmann2024acezero}
Eric Brachmann, Jamie Wynn, Shuai Chen, Tommaso Cavallari, {\'{A}}ron Monszpart, Daniyar Turmukhambetov, and Victor~Adrian Prisacariu.
\newblock Scene coordinate reconstruction: Posing of image collections via incremental learning of a relocalizer.
\newblock In \emph{ECCV}, 2024.

\bibitem[Brahmbhatt et~al.(2018)Brahmbhatt, Gu, Kim, Hays, and Kautz]{Brahmbhatt18}
S. Brahmbhatt, J. Gu, K. Kim, J. Hays, and J. Kautz.
\newblock {Geometry-Aware Learning of Maps for Camera Localization}.
\newblock In \emph{CVPR}, 2018.

\bibitem[Brown and Lowe(2005)]{brown2005unsupervised}
Matthew Brown and David~G. Lowe.
\newblock Unsupervised {3D} object recognition and reconstruction in unordered datasets.
\newblock In \emph{{3DIM}}, 2005.

\bibitem[Cai et~al.(2020)Cai, Yang, Averbuch-Elor, Hao, Belongie, Snavely, and Hariharan]{Ruojin2020ShapeGF}
Ruojin Cai, Guandao Yang, Hadar Averbuch-Elor, Zekun Hao, Serge Belongie, Noah Snavely, and Bharath Hariharan.
\newblock Learning gradient fields for shape generation.
\newblock In \emph{ECCV}, 2020.

\bibitem[Cavallari et~al.(2017)Cavallari, Golodetz, Lord, Valentin, Di~Stefano, and Torr]{cavallari2017fly}
Tommaso Cavallari, Stuart Golodetz, Nicholas~A Lord, Julien Valentin, Luigi Di~Stefano, and Philip~HS Torr.
\newblock On-the-fly adaptation of regression forests for online camera relocalisation.
\newblock In \emph{CVPR}, 2017.

\bibitem[Cavallari et~al.(2019)Cavallari, Bertinetto, Mukhoti, Torr, and Golodetz]{Cavallari2019network}
Tommaso Cavallari, Luca Bertinetto, Jishnu Mukhoti, Philip Torr, and Stuart Golodetz.
\newblock Let's take this online: Adapting scene coordinate regression network predictions for online {RGB-D} camera relocalisation.
\newblock In \emph{3DV}, 2019.

\bibitem[{Cavallari} et~al.(2019){Cavallari}, {Golodetz}, {Lord}, {Valentin}, {Prisacariu}, {Di Stefano}, and {Torr}]{Cavallari2019cascade}
Tommaso {Cavallari}, Stuart {Golodetz}, Nicholas~A. {Lord}, Julien {Valentin}, Victor~A. {Prisacariu}, Luigi {Di Stefano}, and Philip H.~S. {Torr}.
\newblock Real-time {RGB-D} camera pose estimation in novel scenes using a relocalisation cascade.
\newblock \emph{IEEE TPAMI}, 2019.

\bibitem[Chang et~al.(2015)Chang, Funkhouser, Guibas, Hanrahan, Huang, Li, Savarese, Savva, Song, Su, Xiao, Yi, and Yu]{shapenet2015}
Angel~X. Chang, Thomas Funkhouser, Leonidas Guibas, Pat Hanrahan, Qixing Huang, Zimo Li, Silvio Savarese, Manolis Savva, Shuran Song, Hao Su, Jianxiong Xiao, Li Yi, and Fisher Yu.
\newblock {ShapeNet}: An information-rich {3D} model repository.
\newblock \emph{arXiv}, 2015.

\bibitem[Chen et~al.(2021)Chen, Wang, and Prisacariu]{chen21}
Shuai Chen, Zirui Wang, and Victor Prisacariu.
\newblock {Direct-PoseNet}: Absolute pose regression with photometric consistency.
\newblock In \emph{3DV}, 2021.

\bibitem[Chen et~al.(2022)Chen, Li, Wang, and Prisacariu]{chen2022dfnet}
Shuai Chen, Xinghui Li, Zirui Wang, and Victor Prisacariu.
\newblock {DFNet}: Enhance absolute pose regression with direct feature matching.
\newblock In \emph{ECCV}, 2022.

\bibitem[Chen et~al.(2024)Chen, Cavallari, Prisacariu, and Brachmann]{chen2024marepo}
Shuai Chen, Tommaso Cavallari, Victor~Adrian Prisacariu, and Eric Brachmann.
\newblock Map-relative pose regression for visual re-localization.
\newblock In \emph{CVPR}, 2024.

\bibitem[Dai et~al.(2017{\natexlab{a}})Dai, Chang, Savva, Halber, Funkhouser, and Nie{\ss}ner]{dai2017scannet}
Angela Dai, Angel~X Chang, Manolis Savva, Maciej Halber, Thomas Funkhouser, and Matthias Nie{\ss}ner.
\newblock {ScanNet}: Richly-annotated {3D} reconstructions of indoor scenes.
\newblock In \emph{CVPR}, 2017{\natexlab{a}}.

\bibitem[Dai et~al.(2017{\natexlab{b}})Dai, Nie{\ss}ner, Zoll{\"o}fer, Izadi, and Theobalt]{dai2017bundlefusion}
Angela Dai, Matthias Nie{\ss}ner, Michael Zoll{\"o}fer, Shahram Izadi, and Christian Theobalt.
\newblock {BundleFusion}: Real-time globally consistent {3D} reconstruction using on-the-fly surface re-integration.
\newblock \emph{ACM TOG}, 2017{\natexlab{b}}.

\bibitem[Deitke et~al.(2023{\natexlab{a}})Deitke, Liu, Wallingford, Ngo, Michel, Kusupati, Fan, Laforte, Voleti, Gadre, VanderBilt, Kembhavi, Vondrick, Gkioxari, Ehsani, Schmidt, and Farhadi]{objaverseXL}
Matt Deitke, Ruoshi Liu, Matthew Wallingford, Huong Ngo, Oscar Michel, Aditya Kusupati, Alan Fan, Christian Laforte, Vikram Voleti, Samir~Yitzhak Gadre, Eli VanderBilt, Aniruddha Kembhavi, Carl Vondrick, Georgia Gkioxari, Kiana Ehsani, Ludwig Schmidt, and Ali Farhadi.
\newblock {Objaverse-XL}: A universe of {10M+} {3D} objects.
\newblock \emph{arXiv}, 2023{\natexlab{a}}.

\bibitem[Deitke et~al.(2023{\natexlab{b}})Deitke, Schwenk, Salvador, Weihs, Michel, VanderBilt, Schmidt, Ehsani, Kembhavi, and Farhadi]{objaverse}
Matt Deitke, Dustin Schwenk, Jordi Salvador, Luca Weihs, Oscar Michel, Eli VanderBilt, Ludwig Schmidt, Kiana Ehsani, Aniruddha Kembhavi, and Ali Farhadi.
\newblock Objaverse: A universe of annotated {3D} objects.
\newblock In \emph{CVPR}, 2023{\natexlab{b}}.

\bibitem[Dhariwal and Nichol(2021)]{Dhariwal2021guidance}
Prafulla Dhariwal and Alex Nichol.
\newblock Diffusion models beat {GANs} on image synthesis.
\newblock In \emph{NeurIPS}, 2021.

\bibitem[Do et~al.(2022)Do, Miksik, DeGol, Park, and Sinha]{do2022learning}
Tien Do, Ondrej Miksik, Joseph DeGol, Hyun~Soo Park, and Sudipta~N Sinha.
\newblock Learning to detect scene landmarks for camera localization.
\newblock In \emph{CVPR}, 2022.

\bibitem[Dong et~al.(2022)Dong, Wang, Zhuang, Kannala, Pollefeys, and Chen]{dong2022visual}
Siyan Dong, Shuzhe Wang, Yixin Zhuang, Juho Kannala, Marc Pollefeys, and Baoquan Chen.
\newblock Visual localization via few-shot scene region classification.
\newblock In \emph{3DV}, 2022.

\bibitem[Edstedt et~al.(2024)Edstedt, Sun, Bökman, Wadenbäck, and Felsberg]{edstedt2024roma}
Johan Edstedt, Qiyu Sun, Georg Bökman, Mårten Wadenbäck, and Michael Felsberg.
\newblock Roma: Robust dense feature matching.
\newblock In \emph{CVPR}, 2024.

\bibitem[Fischler and Bolles(1981)]{fischler1981random}
Martin~A Fischler and Robert~C Bolles.
\newblock Random sample consensus: a paradigm for model fitting with applications to image analysis and automated cartography.
\newblock \emph{Communications of the ACM}, 1981.

\bibitem[Fu et~al.(2018)Fu, Gong, Wang, Batmanghelich, and Tao]{fu2018deep}
Huan Fu, Mingming Gong, Chaohui Wang, Kayhan Batmanghelich, and Dacheng Tao.
\newblock Deep ordinal regression network for monocular depth estimation.
\newblock In \emph{CVPR}, 2018.

\bibitem[Gao et~al.(2024)Gao, Holynski, Henzler, Brussee, Martin-Brualla, Srinivasan, Barron, and Poole]{gao2024cat3d}
Ruiqi Gao, Aleksander Holynski, Philipp Henzler, Arthur Brussee, Ricardo Martin-Brualla, Pratul~P. Srinivasan, Jonathan~T. Barron, and Ben Poole.
\newblock {CAT3D}: Create anything in {3D} with multi-view diffusion models.
\newblock In \emph{NeurIPS}, 2024.

\bibitem[Gao et~al.(2003)Gao, Hou, Tang, and Cheng]{gao2003complete}
Xiao-Shan Gao, Xiao-Rong Hou, Jianliang Tang, and Hang-Fei Cheng.
\newblock Complete solution classification for the perspective-three-point problem.
\newblock \emph{IEEE TPAMI}, 2003.

\bibitem[Hartley and Zisserman(2003)]{hartley2003multiple}
Richard Hartley and Andrew Zisserman.
\newblock \emph{Multiple view geometry in computer vision}.
\newblock Cambridge University Press, 2003.

\bibitem[He et~al.(2024)He, Sun, Wang, Peng, Huang, Bao, and Zhou]{he2024dfsfm}
Xingyi He, Jiaming Sun, Yifan Wang, Sida Peng, Qixing Huang, Hujun Bao, and Xiaowei Zhou.
\newblock Detector-free structure from motion.
\newblock In \emph{CVPR}, 2024.

\bibitem[Ho and Salimans(2021)]{ho2021classifierfree}
Jonathan Ho and Tim Salimans.
\newblock Classifier-free diffusion guidance.
\newblock In \emph{NeurIPS Workshop}, 2021.

\bibitem[Ho et~al.(2020)Ho, Jain, and Abbeel]{Ho2020DDPM}
Jonathan Ho, Ajay Jain, and Pieter Abbeel.
\newblock Denoising diffusion probabilistic models.
\newblock In \emph{NeurIPS}, 2020.

\bibitem[Humenberger et~al.(2020{\natexlab{a}})Humenberger, Cabon, Guerin, Morat, Leroy, Revaud, Rerole, Pion, de~Souza, and Csurka]{humenberger2020robust}
Martin Humenberger, Yohann Cabon, Nicolas Guerin, Julien Morat, Vincent Leroy, J{\'e}r{\^o}me Revaud, Philippe Rerole, No{\'e} Pion, Cesar de Souza, and Gabriela Csurka.
\newblock Robust image retrieval-based visual localization using kapture.
\newblock \emph{arXiv}, 2020{\natexlab{a}}.

\bibitem[Humenberger et~al.(2020{\natexlab{b}})Humenberger, Cabon, Guerin, Morat, Revaud, Rerole, Pion, de~Souza, Leroy, and Csurka]{kapture2020}
Martin Humenberger, Yohann Cabon, Nicolas Guerin, Julien Morat, Jérôme Revaud, Philippe Rerole, Noé Pion, Cesar de Souza, Vincent Leroy, and Gabriela Csurka.
\newblock Robust image retrieval-based visual localization using {Kapture}.
\newblock \emph{arXiv}, 2020{\natexlab{b}}.

\bibitem[Jin et~al.(2020)Jin, Mishkin, Mishchuk, Matas, Fua, Yi, and Trulls]{Jin2020imc}
Yuhe Jin, Dmytro Mishkin, Anastasiia Mishchuk, Jiri Matas, Pascal Fua, Kwang~Moo Yi, and Eduard Trulls.
\newblock {Image Matching across Wide Baselines: From Paper to Practice}.
\newblock \emph{IJCV}, 2020.

\bibitem[Kendall and Cipolla(2017)]{Kendall17}
A. Kendall and R. Cipolla.
\newblock Geometric loss functions for camera pose regression with deep learning.
\newblock In \emph{CVPR}, 2017.

\bibitem[Kendall et~al.(2015)Kendall, Grimes, and Cipolla]{kendall2015posenet}
Alex Kendall, Matthew Grimes, and Roberto Cipolla.
\newblock Posenet: A convolutional network for real-time 6-{DOF} camera relocalization.
\newblock In \emph{CVPR}, 2015.

\bibitem[Kerbl et~al.(2023)Kerbl, Kopanas, Leimk{\"u}hler, and Drettakis]{kerbl3Dgaussians}
Bernhard Kerbl, Georgios Kopanas, Thomas Leimk{\"u}hler, and George Drettakis.
\newblock {3D} gaussian splatting for real-time radiance field rendering.
\newblock \emph{ACM TOG}, 2023.

\bibitem[Kingma(2014)]{kingma2014adam}
Diederik~P Kingma.
\newblock Adam: A method for stochastic optimization.
\newblock \emph{arXiv}, 2014.

\bibitem[Lee et~al.(2023)Lee, Im, Lee, and Yoon]{Lee2023DiffusionPM}
Jumin Lee, Woobin Im, Sebin Lee, and Sung-Eui Yoon.
\newblock Diffusion probabilistic models for scene-scale {3D} categorical data.
\newblock \emph{arXiv}, 2023.

\bibitem[Leroy et~al.(2024)Leroy, Cabon, and Revaud]{leroy2024mast3r}
Vincent Leroy, Yohann Cabon, and Jerome Revaud.
\newblock Grounding image matching in {3D} with {MASt3R}.
\newblock In \emph{ECCV}, 2024.

\bibitem[Li et~al.(2020)Li, Wang, Zhao, Verbeek, and Kannala]{li2020hscnet}
Xiaotian Li, Shuzhe Wang, Yi Zhao, Jakob Verbeek, and Juho Kannala.
\newblock Hierarchical scene coordinate classification and regression for visual localization.
\newblock In \emph{CVPR}, 2020.

\bibitem[Li and Snavely(2018)]{MegaDepthLi18}
Zhengqi Li and Noah Snavely.
\newblock {MegaDepth}: Learning single-view depth prediction from internet photos.
\newblock In \emph{CVPR}, 2018.

\bibitem[Lindenberger et~al.(2023)Lindenberger, Sarlin, and Pollefeys]{lindenberger2023lightglue}
Philipp Lindenberger, Paul-Edouard Sarlin, and Marc Pollefeys.
\newblock {LightGlue}: Local feature matching at light speed.
\newblock In \emph{ICCV}, 2023.

\bibitem[Liu et~al.(2015)Liu, Shen, Lin, and Reid]{liu2015learning}
Fayao Liu, Chunhua Shen, Guosheng Lin, and Ian Reid.
\newblock Learning depth from single monocular images using deep convolutional neural fields.
\newblock \emph{IEEE TPAMI}, 2015.

\bibitem[Liu et~al.(2025)Liu, Yang, Liu, Huang, Chiang, Kong, Kobori, and Lee]{liu2025reprojection}
Ting-Ru Liu, Hsuan-Kung Yang, Jou-Min Liu, Chun-Wei Huang, Tsung-Chih Chiang, Quan Kong, Norimasa Kobori, and Chun-Yi Lee.
\newblock Reprojection errors as prompts for efficient scene coordinate regression.
\newblock In \emph{ECCV}, 2025.

\bibitem[Liu et~al.(2024)Liu, Li, Li, Qi, Li, and Yang]{liu2024pyramiddiffusionfine3d}
Yuheng Liu, Xinke Li, Xueting Li, Lu Qi, Chongshou Li, and Ming-Hsuan Yang.
\newblock Pyramid diffusion for fine {3D} large scene generation.
\newblock In \emph{ECCV}, 2024.

\bibitem[Liu et~al.(2019{\natexlab{a}})Liu, Tang, Lin, and Han]{liu2019point}
Zhijian Liu, Haotian Tang, Yujun Lin, and Song Han.
\newblock Point-voxel cnn for efficient {3D} deep learning.
\newblock \emph{NeurIPS}, 2019{\natexlab{a}}.

\bibitem[Liu et~al.(2019{\natexlab{b}})Liu, Tang, Lin, and Han]{liu2019pvcnn}
Zhijian Liu, Haotian Tang, Yujun Lin, and Song Han.
\newblock Point-voxel {CNN} for efficient {3D} deep learning.
\newblock In \emph{NeurIPS}, 2019{\natexlab{b}}.

\bibitem[Luo(2022)]{Luo2022UnderstandingDM}
Calvin Luo.
\newblock Understanding diffusion models: A unified perspective.
\newblock \emph{arXiv}, 2022.

\bibitem[Luo and Hu(2021)]{luo2021diffusion}
Shitong Luo and Wei Hu.
\newblock Diffusion probabilistic models for {3D} point cloud generation.
\newblock In \emph{CVPR}, 2021.

\bibitem[Lyu et~al.(2022)Lyu, Kong, Xu, Pan, and Lin]{Lyu2022diffusionrefinement}
Zhaoyang Lyu, Zhifeng Kong, Xudong Xu, Liang Pan, and Dahua Lin.
\newblock A conditional point diffusion-refinement paradigm for {3D} point cloud completion.
\newblock In \emph{ICLR}, 2022.

\bibitem[Melas-Kyriazi et~al.(2023)Melas-Kyriazi, Rupprecht, and Vedaldi]{melaskyriazi2023pc2}
Luke Melas-Kyriazi, Christian Rupprecht, and Andrea Vedaldi.
\newblock {PC2}: Projection-conditioned point cloud diffusion for single-image {3D} reconstruction.
\newblock In \emph{CVPR}, 2023.

\bibitem[Mildenhall et~al.(2020)Mildenhall, Srinivasan, Tancik, Barron, Ramamoorthi, and Ng]{mildenhall2020nerf}
Ben Mildenhall, Pratul~P. Srinivasan, Matthew Tancik, Jonathan~T. Barron, Ravi Ramamoorthi, and Ren Ng.
\newblock {NeRF}: Representing scenes as neural radiance fields for view synthesis.
\newblock In \emph{ECCV}, 2020.

\bibitem[Mo et~al.(2023)Mo, Xie, Chu, HONG, Nie{\ss}ner, and Li]{mo2023ditd}
Shentong Mo, Enze Xie, Ruihang Chu, Lanqing HONG, Matthias Nie{\ss}ner, and Zhenguo Li.
\newblock {DiT-3D}: Exploring plain diffusion transformers for {3D} shape generation.
\newblock In \emph{NeurIPS}, 2023.

\bibitem[Moreau et~al.(2021)Moreau, Piasco, Tsishkou, Stanciulescu, and de~La~Fortelle]{Moreau21lens}
Arthur Moreau, Nathan Piasco, Dzmitry Tsishkou, Bogdan Stanciulescu, and Arnaud de La~Fortelle.
\newblock {LENS}: Localization enhanced by {NeRF} synthesis.
\newblock In \emph{CoRL}, 2021.

\bibitem[Nguyen et~al.(2024)Nguyen, Fontan, Milford, and Fischer]{nguyen2024focustune}
Son~Tung Nguyen, Alejandro Fontan, Michael Milford, and Tobias Fischer.
\newblock Focustune: Tuning visual localization through focus-guided sampling.
\newblock In \emph{WACV}, 2024.

\bibitem[Nguyen et~al.(2025)Nguyen, Piasco, Luis~Roldão, Tsishkou, Caraffa, Tarel, and Brémond]{nguyen2025pointmapdiff}
Thang-Anh-Quan Nguyen, Nathan Piasco, Moussab~Bennehar Luis~Roldão, Dzmitry Tsishkou, Laurent Caraffa, Jean-Philippe Tarel, and Roland Brémond.
\newblock Pointmap-conditioned diffusion for consistent novel view synthesis.
\newblock \emph{arXiv}, 2025.

\bibitem[Niemeyer et~al.(2022)Niemeyer, Barron, Mildenhall, Sajjadi, Geiger, and Radwan]{Niemeyer2021Regnerf}
Michael Niemeyer, Jonathan~T. Barron, Ben Mildenhall, Mehdi S.~M. Sajjadi, Andreas Geiger, and Noha Radwan.
\newblock {RegNeRF}: Regularizing neural radiance fields for view synthesis from sparse inputs.
\newblock In \emph{CVPR}, 2022.

\bibitem[Nunes et~al.(2024)Nunes, Marcuzzi, Mersch, Behley, and Stachniss]{nunes2024lidarcompletion}
Lucas Nunes, Rodrigo Marcuzzi, Benedikt Mersch, Jens Behley, and Cyrill Stachniss.
\newblock Scaling diffusion models to real-world {3D} {LiDAR} scene completion.
\newblock In \emph{CVPR}, 2024.

\bibitem[Pan et~al.(2024)Pan, Baráth, Pollefeys, and Sch\"{o}nberger]{pan2024glomap}
Linfei Pan, Dániel Baráth, Marc Pollefeys, and Johannes~Lutz Sch\"{o}nberger.
\newblock Global structure-from-motion revisited.
\newblock In \emph{ECCV}, 2024.

\bibitem[Paszke et~al.(2017)Paszke, Gross, Chintala, Chanan, Yang, DeVito, Lin, Desmaison, Antiga, and Lerer]{paszke2017automatic}
Adam Paszke, Sam Gross, Soumith Chintala, Gregory Chanan, Edward Yang, Zachary DeVito, Zeming Lin, Alban Desmaison, Luca Antiga, and Adam Lerer.
\newblock Automatic differentiation in {PyTorch}.
\newblock In \emph{NeurIPS Workshop}, 2017.

\bibitem[Poole et~al.(2023)Poole, Jain, Barron, and Mildenhall]{poole2023dreamfusion}
Ben Poole, Ajay Jain, Jonathan~T. Barron, and Ben Mildenhall.
\newblock {DreamFusion}: Text-to-{3D} using {2D} diffusion.
\newblock In \emph{ICLR}, 2023.

\bibitem[Qi et~al.(2017{\natexlab{a}})Qi, Su, Mo, and Guibas]{qi2016pointnet}
Charles~R. Qi, Hao Su, Kaichun Mo, and Leonidas~J. Guibas.
\newblock {PointNet}: Deep learning on point sets for {3D} classification and segmentation.
\newblock In \emph{CVPR}, 2017{\natexlab{a}}.

\bibitem[Qi et~al.(2017{\natexlab{b}})Qi, Yi, Su, and Guibas]{qi2017pointnetplusplus}
Charles~R Qi, Li Yi, Hao Su, and Leonidas~J Guibas.
\newblock {PointNet++}: Deep hierarchical feature learning on point sets in a metric space.
\newblock In \emph{NeurIPS}, 2017{\natexlab{b}}.

\bibitem[Ran et~al.(2024)Ran, Guizilini, and Wang]{ran2024towards}
Haoxi Ran, Vitor Guizilini, and Yue Wang.
\newblock Towards realistic scene generation with {LiDAR} diffusion models.
\newblock In \emph{CVPR}, 2024.

\bibitem[Ranftl et~al.(2021)Ranftl, Bochkovskiy, and Koltun]{ranftl2021vision}
Ren{\'e} Ranftl, Alexey Bochkovskiy, and Vladlen Koltun.
\newblock Vision transformers for dense prediction.
\newblock In \emph{ICCV}, 2021.

\bibitem[Reizenstein et~al.(2021)Reizenstein, Shapovalov, Henzler, Sbordone, Labatut, and Novotny]{reizenstein21co3d}
Jeremy Reizenstein, Roman Shapovalov, Philipp Henzler, Luca Sbordone, Patrick Labatut, and David Novotny.
\newblock Common objects in {3D}: Large-scale learning and evaluation of real-life {3D} category reconstruction.
\newblock In \emph{ICCV}, 2021.

\bibitem[Revaud et~al.(2024)Revaud, Cabon, Brégier, Lee, and Weinzaepfel]{Revaud2024sacreg}
Jerome Revaud, Yohann Cabon, Romain Brégier, JongMin Lee, and Philippe Weinzaepfel.
\newblock {SACReg}: Scene-agnostic coordinate regression for visual localization.
\newblock In \emph{CVPRW}, 2024.

\bibitem[Roessle et~al.(2022)Roessle, Barron, Mildenhall, Srinivasan, and Nie{\ss}ner]{roessle2022depthpriorsnerf}
Barbara Roessle, Jonathan~T. Barron, Ben Mildenhall, Pratul~P. Srinivasan, and Matthias Nie{\ss}ner.
\newblock Dense depth priors for neural radiance fields from sparse input views.
\newblock In \emph{CVPR}, 2022.

\bibitem[Sarlin et~al.(2019)Sarlin, Cadena, Siegwart, and Dymczyk]{sarlin2019coarse}
Paul-Edouard Sarlin, Cesar Cadena, Roland Siegwart, and Marcin Dymczyk.
\newblock From coarse to fine: Robust hierarchical localization at large scale.
\newblock In \emph{CVPR}, 2019.

\bibitem[Sarlin et~al.(2020)Sarlin, DeTone, Malisiewicz, and Rabinovich]{sarlin20superglue}
Paul-Edouard Sarlin, Daniel DeTone, Tomasz Malisiewicz, and Andrew Rabinovich.
\newblock {SuperGlue}: Learning feature matching with graph neural networks.
\newblock In \emph{CVPR}, 2020.

\bibitem[Sarlin et~al.(2021)Sarlin, Unagar, Larsson, Germain, Toft, Larsson, Pollefeys, Lepetit, Hammarstrand, Kahl, et~al.]{sarlin2021back}
Paul-Edouard Sarlin, Ajaykumar Unagar, Mans Larsson, Hugo Germain, Carl Toft, Viktor Larsson, Marc Pollefeys, Vincent Lepetit, Lars Hammarstrand, Fredrik Kahl, et~al.
\newblock Back to the feature: Learning robust camera localization from pixels to pose.
\newblock In \emph{CVPR}, 2021.

\bibitem[Sattler et~al.(2011)Sattler, Leibe, and Kobbelt]{Sattler11}
T. Sattler, B. Leibe, and L. Kobbelt.
\newblock Fast image-based localization using direct {2D-to-3D} matching.
\newblock In \emph{ICCV}, 2011.

\bibitem[Sattler et~al.(2016)Sattler, Leibe, and Kobbelt]{sattler2016efficient}
Torsten Sattler, Bastian Leibe, and Leif Kobbelt.
\newblock Efficient \& effective prioritized matching for large-scale image-based localization.
\newblock \emph{PAMI}, 2016.

\bibitem[Sattler et~al.(2017)Sattler, Leibe, and Kobbelt]{Sattler2017AS}
Torsten Sattler, Bastian Leibe, and Leif Kobbelt.
\newblock Efficient \& effective prioritized matching for large-scale image-based localization.
\newblock \emph{IEEE TPAMI}, 2017.

\bibitem[Sattler et~al.(2019)Sattler, Zhou, Pollefeys, and Leal-Taixe]{sattler2019limits}
Torsten Sattler, Qunjie Zhou, Marc Pollefeys, and Laura Leal-Taixe.
\newblock Understanding the limitations of cnn-based absolute camera pose regression.
\newblock In \emph{CVPR}, 2019.

\bibitem[Schonberger and Frahm(2016)]{schonberger2016structure}
Johannes~L Schonberger and Jan-Michael Frahm.
\newblock Structure-from-motion revisited.
\newblock In \emph{CVPR}, 2016.

\bibitem[Shavit et~al.(2021)Shavit, Ferens, and Keller]{Shavit21multiscene}
Yoli Shavit, Ron Ferens, and Yosi Keller.
\newblock Learning multi-scene absolute pose regression with transformers.
\newblock In \emph{ICCV}, 2021.

\bibitem[Shotton et~al.(2013)Shotton, Glocker, Zach, Izadi, Criminisi, and Fitzgibbon]{shotton2013scene}
Jamie Shotton, Ben Glocker, Christopher Zach, Shahram Izadi, Antonio Criminisi, and Andrew Fitzgibbon.
\newblock Scene coordinate regression forests for camera relocalization in {RGB-D} images.
\newblock In \emph{CVPR}, 2013.

\bibitem[Snavely et~al.(2006)Snavely, Seitz, and Szeliski]{Snavely2006photo}
Noah Snavely, Steven Seitz, and Richard Szeliski.
\newblock Photo tourism: exploring photo collections in {3D}.
\newblock \emph{ACM TOG}, 2006.

\bibitem[Sohl-Dickstein et~al.(2015)Sohl-Dickstein, Weiss, Maheswaranathan, and Ganguli]{SohlDickstein2015diffusion}
Jascha Sohl-Dickstein, Eric Weiss, Niru Maheswaranathan, and Surya Ganguli.
\newblock Deep unsupervised learning using nonequilibrium thermodynamics.
\newblock In \emph{ICML}, 2015.

\bibitem[Song et~al.(2021{\natexlab{a}})Song, Meng, and Ermon]{song2021ddim}
Jiaming Song, Chenlin Meng, and Stefano Ermon.
\newblock Denoising diffusion implicit models.
\newblock In \emph{ICLR}, 2021{\natexlab{a}}.

\bibitem[Song and Ermon(2019)]{yang2019datagradients}
Yang Song and Stefano Ermon.
\newblock Generative modeling by estimating gradients of the data distribution.
\newblock In \emph{NeurIPS}, 2019.

\bibitem[Song et~al.(2021{\natexlab{b}})Song, Sohl-Dickstein, Kingma, Kumar, Ermon, and Poole]{song2021scorebased}
Yang Song, Jascha Sohl-Dickstein, Diederik~P Kingma, Abhishek Kumar, Stefano Ermon, and Ben Poole.
\newblock Score-based generative modeling through stochastic differential equations.
\newblock In \emph{ICLR}, 2021{\natexlab{b}}.

\bibitem[Sturm et~al.(2012)Sturm, Engelhard, Endres, Burgard, and Cremers]{sturm12iros}
J. Sturm, N. Engelhard, F. Endres, W. Burgard, and D. Cremers.
\newblock A benchmark for the evaluation of {RGB-D SLAM} systems.
\newblock In \emph{IROS}, 2012.

\bibitem[Sun et~al.(2021)Sun, Shen, Wang, Bao, and Zhou]{sun2021loftr}
Jiaming Sun, Zehong Shen, Yuang Wang, Hujun Bao, and Xiaowei Zhou.
\newblock {LoFTR}: Detector-free local feature matching with transformers.
\newblock In \emph{CVPR}, 2021.

\bibitem[Tancik et~al.(2023)Tancik, Weber, Ng, Li, Yi, Kerr, Wang, Kristoffersen, Austin, Salahi, Ahuja, McAllister, and Kanazawa]{nerfstudio}
Matthew Tancik, Ethan Weber, Evonne Ng, Ruilong Li, Brent Yi, Justin Kerr, Terrance Wang, Alexander Kristoffersen, Jake Austin, Kamyar Salahi, Abhik Ahuja, David McAllister, and Angjoo Kanazawa.
\newblock Nerfstudio: A modular framework for neural radiance field development.
\newblock In \emph{SIGGRAPH}, 2023.

\bibitem[Vincent(2011)]{vincent2011connection}
Pascal Vincent.
\newblock A connection between score matching and denoising autoencoders.
\newblock \emph{Neural computation}, 2011.

\bibitem[Waechter et~al.(2017)Waechter, Beljan, Fuhrmann, Moehrle, Kopf, and Goesele]{waechter2017rephotography}
Michael Waechter, Mate Beljan, Simon Fuhrmann, Nils Moehrle, Johannes Kopf, and Michael Goesele.
\newblock Virtual rephotography: Novel view prediction error for {3D} reconstruction.
\newblock \emph{ACM TOG}, 2017.

\bibitem[Wang et~al.(2024{\natexlab{a}})Wang, Jiang, Galliani, Vogel, and Pollefeys]{wang2024glace}
Fangjinhua Wang, Xudong Jiang, Silvano Galliani, Christoph Vogel, and Marc Pollefeys.
\newblock {GLACE}: Global local accelerated coordinate encoding.
\newblock In \emph{CVPR}, 2024{\natexlab{a}}.

\bibitem[Wang et~al.(2024{\natexlab{b}})Wang, Karaev, Rupprecht, and Novotny]{wang2024vggsfm}
Jianyuan Wang, Nikita Karaev, Christian Rupprecht, and David Novotny.
\newblock {VGGSfM}: Visual geometry grounded deep structure from motion.
\newblock In \emph{CVPR}, 2024{\natexlab{b}}.

\bibitem[Wang et~al.(2024{\natexlab{c}})Wang, Laskar, Melekhov, Li, Zhao, Tolias, and Kannala]{Wang2024hscnetpp}
Shuzhe Wang, Zakaria Laskar, Iaroslav Melekhov, Xiaotian Li, Yi Zhao, Giorgos Tolias, and Juho Kannala.
\newblock {HSCNet++} : Hierarchical scene coordinate classification and regression for visual localization with transformer.
\newblock \emph{IJCV}, 2024{\natexlab{c}}.

\bibitem[Wang et~al.(2024{\natexlab{d}})Wang, Leroy, Cabon, Chidlovskii, and Revaud]{wang2024dust3r}
Shuzhe Wang, Vincent Leroy, Yohann Cabon, Boris Chidlovskii, and Jerome Revaud.
\newblock {DUSt3R}: Geometric {3D} vision made easy.
\newblock In \emph{CVPR}, 2024{\natexlab{d}}.

\bibitem[Wu(2013)]{Wu2013visualSfM}
Changchang Wu.
\newblock Towards linear-time incremental structure from motion.
\newblock In \emph{3DV}, 2013.

\bibitem[Wu et~al.(2024)Wu, Mildenhall, Henzler, Park, Gao, Watson, Srinivasan, Verbin, Barron, Poole, and Holynski]{wu2024reconfusion}
Rundi Wu, Ben Mildenhall, Philipp Henzler, Keunhong Park, Ruiqi Gao, Daniel Watson, Pratul~P. Srinivasan, Dor Verbin, Jonathan~T. Barron, Ben Poole, and Aleksander Holynski.
\newblock {ReconFusion}: {3D} reconstruction with diffusion priors.
\newblock In \emph{CVPR}, 2024.

\bibitem[Wynn and Turmukhambetov(2023)]{wynn2023diffusionerf}
Jamie Wynn and Daniyar Turmukhambetov.
\newblock {DiffusioNeRF}: Regularizing neural radiance fields with denoising diffusion models.
\newblock In \emph{CVPR}, 2023.

\bibitem[Xiong et~al.(2023)Xiong, Muttukuru, Upadhyay, Chari, and Kadambi]{xiong2023sparsegs}
Haolin Xiong, Sairisheek Muttukuru, Rishi Upadhyay, Pradyumna Chari, and Achuta Kadambi.
\newblock {SparseGS}: Real-time 360° sparse view synthesis using gaussian splatting.
\newblock \emph{arXiv}, 2023.

\bibitem[Yang et~al.(2019{\natexlab{a}})Yang, Huang, Hao, Liu, Belongie, and Hariharan]{Yang2019pointflow}
Guandao Yang, Xun Huang, Zekun Hao, Ming-Yu Liu, Serge Belongie, and Bharath Hariharan.
\newblock {PointFlow}: {3D} point cloud generation with continuous normalizing flows.
\newblock In \emph{ICCV}, 2019{\natexlab{a}}.

\bibitem[Yang et~al.(2019{\natexlab{b}})Yang, Bai, Tang, Li, Furukawa, and Tan]{Luwei2019sanet}
Luwei Yang, Ziqian Bai, Chengzhou Tang, Honghua Li, Yasutaka Furukawa, and Ping Tan.
\newblock {SANet}: Scene agnostic network for camera localization.
\newblock In \emph{ICCV}, 2019{\natexlab{b}}.

\bibitem[Ye et~al.(2024)Ye, Li, Kerr, Turkulainen, Yi, Pan, Seiskari, Ye, Hu, Tancik, and Kanazawa]{ye2024gsplat}
Vickie Ye, Ruilong Li, Justin Kerr, Matias Turkulainen, Brent Yi, Zhuoyang Pan, Otto Seiskari, Jianbo Ye, Jeffrey Hu, Matthew Tancik, and Angjoo Kanazawa.
\newblock gsplat: An open-source library for {Gaussian} splatting.
\newblock \emph{arXiv}, 2024.

\bibitem[Yeshwanth et~al.(2023)Yeshwanth, Liu, Nie{\ss}ner, and Dai]{yeshwanthliu2023scannetpp}
Chandan Yeshwanth, Yueh-Cheng Liu, Matthias Nie{\ss}ner, and Angela Dai.
\newblock {ScanNet++}: A high-fidelity dataset of {3D} indoor scenes.
\newblock In \emph{ICCV}, 2023.

\bibitem[Zeng et~al.(2022)Zeng, Vahdat, Williams, Gojcic, Litany, Fidler, and Kreis]{zeng2022lion}
Xiaohui Zeng, Arash Vahdat, Francis Williams, Zan Gojcic, Or Litany, Sanja Fidler, and Karsten Kreis.
\newblock {LION}: Latent point diffusion models for {3D} shape generation.
\newblock In \emph{NeurIPS}, 2022.

\bibitem[Zhou et~al.(2021)Zhou, Du, and Wu]{Zhou2021PVD}
Linqi Zhou, Yilun Du, and Jiajun Wu.
\newblock {3D} shape generation and completion through point-voxel diffusion.
\newblock In \emph{ICCV}, 2021.

\bibitem[Zyrianov et~al.(2022)Zyrianov, Zhu, and Wang]{zyrianov2022learning}
Vlas Zyrianov, Xiyue Zhu, and Shenlong Wang.
\newblock Learning to generate realistic {LiDAR} point cloud.
\newblock In \emph{ECCV}, 2022.

\end{thebibliography}
